\documentclass{article} 
\usepackage{iclr2025_conference,times}

\usepackage{amsmath,amsfonts,bm}

\def\eqref#1{equation~\ref{#1}}

\def\1{\bm{1}}

\DeclareMathAlphabet{\mathsfit}{\encodingdefault}{\sfdefault}{m}{sl}
\SetMathAlphabet{\mathsfit}{bold}{\encodingdefault}{\sfdefault}{bx}{n}

\usepackage[T1]{fontenc}
\usepackage[utf8]{inputenc}

\usepackage{url} 
\usepackage{booktabs}
\usepackage{xcolor}
\usepackage{colortbl}
\usepackage{times}
\usepackage{latexsym}
\usepackage{makecell}
\usepackage{lipsum}
\usepackage[toc,page,header]{appendix}
\usepackage{minitoc}

\usepackage{graphicx}
\usepackage{tabularx}
\usepackage{amssymb}
\usepackage{microtype}
\usepackage{wrapfig}
\usepackage{hyperref}
\usepackage{inconsolata}
\usepackage{amsmath}
\usepackage{multirow}
\usepackage{float}
\usepackage[capitalize,noabbrev]{cleveref}
\usepackage{titletoc}
\usepackage{listings}
\usepackage{amsmath,amssymb}
\usepackage[most]{tcolorbox}
\usepackage{algorithm}
\usepackage{algpseudocode}
\usepackage{enumitem}

\usepackage{tikz}
\usepackage{amsmath}
\usepackage{amssymb}
\usepackage{array}
\usepackage{booktabs}
\usepackage{pifont}
\usepackage{xspace}

\newcommand{\kgw}{\textsc{KGW}\xspace}
\newcommand{\aar}{\textsc{Aar}\xspace}
{\begin{itemize}[itemsep=0pt, topsep=0pt]}%
{\end{itemize}}

\title{Mark Your LLM: Detecting the Misuse of Open-Source Large Language Models via Watermarking}

\renewcommand{\thefootnote}{\fnsymbol{footnote}}

\author{
  Yijie Xu\textsuperscript{1,\footnotemark[2]}, 
  Aiwei Liu\textsuperscript{2,\footnotemark[2]}, 
  Xuming Hu\textsuperscript{1,3,\footnotemark[1] \ }, 
  Lijie Wen\textsuperscript{2}, 
  Hui Xiong\textsuperscript{1,3,\footnotemark[1] \ }\\
  \textsuperscript{1}The Hong Kong University of Science and Technology (Guangzhou), Guangzhou, China\\
  \textsuperscript{2}Tsinghua University, Beijing, China\\
  \textsuperscript{3}The Hong Kong University of Science and Technology, Hong Kong SAR, China\\
    \texttt{\href{mailto:yxu409@connect.hkust-gz.edu.cn}{yxu409@connect.hkust-gz.edu.cn}},\ 
    \texttt{\href{mailto:liuaw20@mails.tsinghua.edu.cn}{liuaw20@mails.tsinghua.edu.cn}} \\
    \texttt{\href{mailto:xuminghu@hkust-gz.edu.cn}{xuminghu@hkust-gz.edu.cn}},\ 
    \texttt{\href{mailto:wenlj@tsinghua.edu.cn}{wenlj@tsinghua.edu.cn}},\ 
    \texttt{\href{mailto:xionghui@ust.hk}{xionghui@ust.hk}}
}

\iclrfinalcopy 
\begin{document}
\doparttoc 
\faketableofcontents 

\maketitle
\footnotetext[2]{Equal contribution; more junior authors listed earlier.}
\footnotetext[1]{Corresponding authors.}

\renewcommand{\thefootnote}{\arabic{footnote}}

\begin{abstract}
As open-source large language models (LLMs) like Llama3 become more capable, it is crucial to develop watermarking techniques to detect their potential misuse. Existing watermarking methods either add watermarks during LLM inference, which is unsuitable for open-source LLMs, or primarily target classification LLMs rather than recent generative LLMs. Adapting these watermarks to open-source LLMs for misuse detection remains an open challenge. This work defines two misuse scenarios for open-source LLMs: \textit{Intellectual Property (IP) violation} and \textit{LLM Usage Violation}. Then, we explore the application of inference-time watermark distillation and backdoor watermarking in these contexts. We propose comprehensive evaluation methods to assess the impact of various real-world continual fine-tuning scenarios on watermarks and the effect of these watermarks on LLM performance. Our experiments reveal that backdoor watermarking could effectively detect IP Violation, while inference-time watermark distillation is applicable in both scenarios but less robust to continual fine-tuning and has a more significant impact on LLM performance compared to backdoor watermarking. Exploring more advanced watermarking methods for open-source LLMs to detect their misuse should be an important future direction.
\end{abstract}

\section{Introduction}

With the significant advancements in open-source Large Language Models (LLMs) like Llama3~\cite{dubey2024llama} and Deepseek series~\cite{liu2024deepseek, guo2025deepseek} in terms of reasoning~\cite{qiao-etal-2023-reasoning}, generation~\cite{10.1145/3649449}, and instruction-following~\cite{zeng2023evaluating} capabilities, developers and enterprises can leverage the power of LLMs for downstream tasks more conveniently, such as retrieval~\cite{daiseper, dai2024improve, qiu2024entropy, li2024refiner}, recommendation~\cite{qiu2024ease}, graph-text information fusion~\cite{chen2024deep, chen2024shopping}, and multi-modal tasks~\cite{longvideohaystack, yan2024survey}. Under this context, the misuse of open-source LLMs has become an urgent topic. It primarily involves the theft of LLM intellectual property rights~\cite{ren2024copyright} and using LLMs to generate harmful content for online dissemination~\cite{chen2023combating}.

LLM watermarking techniques~\cite{liu2024survey} are considered an effective method to detect the misuse of LLMs. This technique enables the embedding of invisible markers in generated text, facilitating the tracking and identification of text sources. However, mainstream LLM watermarking techniques primarily rely on inference-time methods that modify output probabilities to add watermarks~\cite{kirchenbauer2023watermark, kuditipudi2023robust}.
These post-processing watermarking algorithms are not applicable to open-source LLMs, as open-source users can easily remove such watermarking processing codes. Watermarking techniques must have sufficient imperceptibility for open-source LLMs, meaning the watermark needs to be embedded into the LLM's parameters.

Currently, some research has begun to explore ways to integrate watermarking algorithms into LLM parameters~\cite{pan2024markllm}. These algorithms include distilling the features of inference-time watermarking algorithms into LLM parameters~\cite{gu2023learnability} and backdoor-based watermarking algorithms~\cite{xu2023instructions} that exhibit watermark features under specific trigger conditions, which are more commonly applied to classification LLMs than generative LLMs. While these algorithms have shown some potential for open-source LLMs~\cite{pan2025can, liu2024can}, how to adapt them to detect misuse of open-source LLMs in real scenarios is still lacking discussion.

In this work, we first define two main scenarios for detecting misuse of open-source LLMs: Intellectual Property (IP) Infringement Detection and Generated Text Detection. We then introduce how to apply watermarking algorithms to these two scenarios, adapting both backdoor watermarking and inference-time watermark distillation to the scenarios. Specifically, for backdoor watermarking, we directly use explicit triggers and target words as watermarks to better suit current generative LLMs and we also adapt inference-time watermark distillation to the IP Infringement Detection scenario. 

\begin{wrapfigure}{r}{0.4\linewidth}
    \vspace{-1mm}
    \centering
    \includegraphics[width=\linewidth]{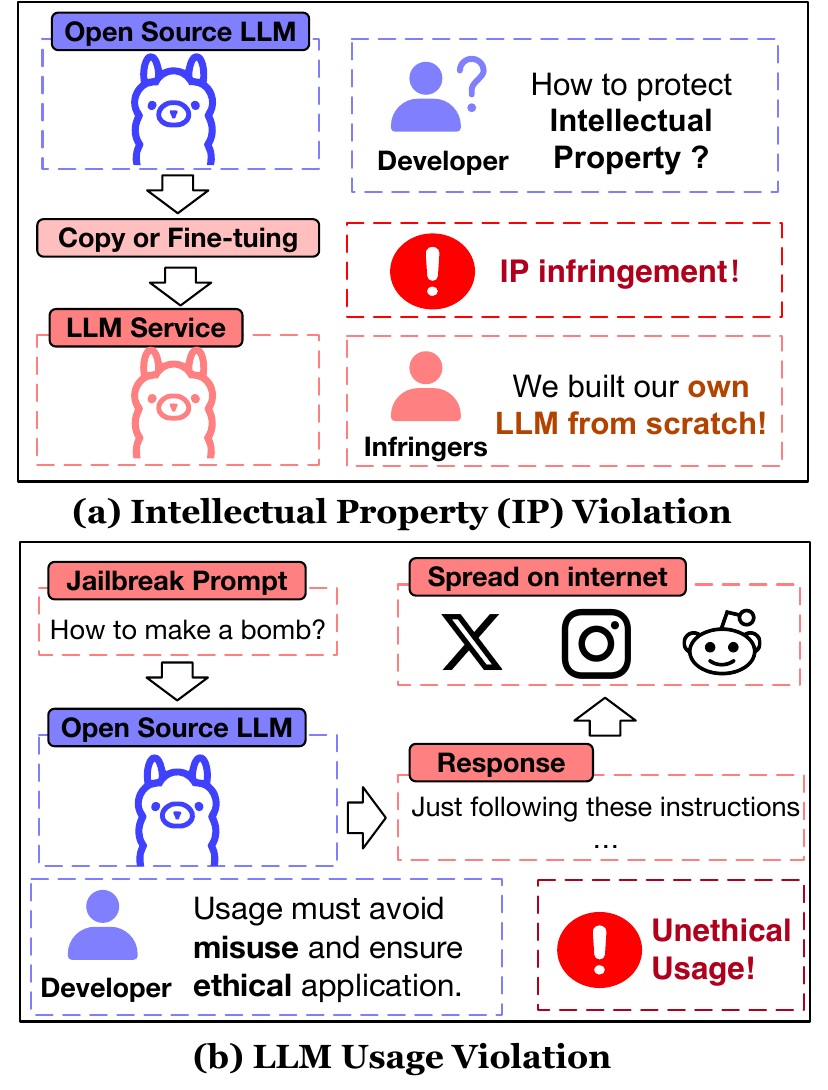}
    \vspace{-5mm}
    \caption{The two main misuse scenarios for LLMs in this work: Intellectual Property Violation (\cref{sec:scenarios-model}) and LLM Usage Violation (\cref{sec:scenarios-text}).}
    \label{fig:intro}
\vspace{-4mm}
\end{wrapfigure}

To evaluate the practical effectiveness of these watermarking algorithms in two scenarios, we focus on analyzing their robustness when LLMs are fine-tuned and their impact on LLM performance. Regarding the robustness of further fine-tuning, we fully consider various scenarios where users fine-tune open-source LLMs, including further pretraining (PT), instruction tuning (IT), DPO~\cite{rafailov2023direct, liu2024tis} or RLHF~\cite{ouyang2022training} for preference optimization. We also consider full-parameter fine-tuning and low-resource fine-tuning, such as LoRA~\cite{hu2021lora}. Regarding the impact of watermarking on LLM performance, we comprehensively evaluated reasoning, understanding, and generation capabilities, assessing reasoning and understanding abilities on datasets like ARC-Challenge~\cite{clark2018think}, MMLU~\cite{hendrycks2020measuring}, and HellaSwag~\cite{zellers2019hellaswag}, and evaluating perplexity (PPL) and proportion of repetitions in generated text on WikiText dataset\cite{merity2016pointer}.

In the experiments, we found that backdoor-based watermarks are a good solution for the intellectual property detection scenario, as they are highly robust to various fine-tuning processes and have minimal impact on LLM performance. However, they cannot address the output text detection scenario. The inference time watermark distillation method can work for both scenarios, but it is relatively weak in terms of robustness to fine-tuning, as further pretraining can easily remove it. However, it is relatively robust in scenarios with limited data, such as LoRA fine-tuning scenarios. Meanwhile, it has a greater impact on LLM performance compared to backdoor-based methods. Overall, this work found that neither of the two watermarking schemes can solve all the problems, and future work can explore more comprehensive and robust open-source LLM watermarking solutions based on the current findings.

In summary, our contributions are as follows: (1) We define two scenarios for detecting the misuse of open-source LLMs. (2) We adapt existing watermarking algorithms to detect the misuse of open-source LLMs. (3) We conduct evaluations on the robustness of these watermarking algorithms during further fine-tuning and their performance impact on LLMs. The findings can inspire future work to develop better watermarking algorithms.

\vspace{-0.5em}
\section{Related Work}
\vspace{-0.5em}
Inference time watermark and backdoor watermark are the main watermarking methods for LLMs~\cite{liu2024survey}, but both have limitations in detecting misuse of open-source LLMs.

\noindent \textbf{Inference time watermark} refers to embedding a watermark by introducing small biases~\cite{kirchenbauer2023watermark} in the logits or by adjusting token sampling preferences~\cite{kuditipudi2023robust}. Despite various optimizations, such as improving robustness to watermarked text modification~\cite{he2024can, liu2023unforgeable, zhao2023provable, liu2023semantic}, minimizing quality impact~\cite{hu2023unbiased}, supporting public detection~\cite{liu2023unforgeable}, and detecting in low-entropy environments~\cite{pan2024waterseeker, lu2024entropy, lee2023wrote}, these watermarks are added post-generation and are thus unsuitable for open-source LLMs.~\citet{gu2023learnability} attempted to have LLMs learn to generate outputs with such watermarks during training, progress made, but the practical application and evaluation in detecting the misuse of open-source LLMs remain limited.

\noindent \textbf{Embedding backdoor watermarks} in LLMs implies that the model generates predefined outputs when encountering specific trigger words or features. This technique has great potential for protecting LLM copyrights. However, previous backdoor watermarks have been typically limited to classification tasks~\cite{liu2021removing, shafieinejad2021robustness}, with limited adaptation for generative LLMs. Although ~\citet{xu2023instructions} explored backdoor attacks during instruction tuning for generative LLMs, their testing was confined to sentiment classification. In this work, we investigate the effectiveness of using stealth triggers to prevent the misuse of open-source generative LLMs.

\noindent \textbf{Fingerprinting} in LLMs involves embedding identifiable signatures or patterns within model parameters, allowing attribution of generated content and verification of model ownership. Output-based techniques~\cite{jin2024proflingo, russinovich2024hey} use adversarial prompts and cryptographic hashing to verify ownership in black-box settings.~\cite{iourovitski2024hide} refine prompts through evolutionary learning. Lightweight methods~\cite{xu2024fp} introduce vector addition for efficient fingerprinting, while instructional fingerprinting~\cite{xu2024instructional} uses backdoor-based tuning. Weight-level methods, such as quantization-based watermarking~\cite{li2023watermarking}, embed identifiers directly into model weights to ensure tamper resistance. Finally, black-box vector-space analysis~\cite{yang2024fingerprint} models unique output spaces for infringement detection in fine-tuning scenarios.

\begin{figure*}
    \centering
    \includegraphics[width=1 \textwidth]{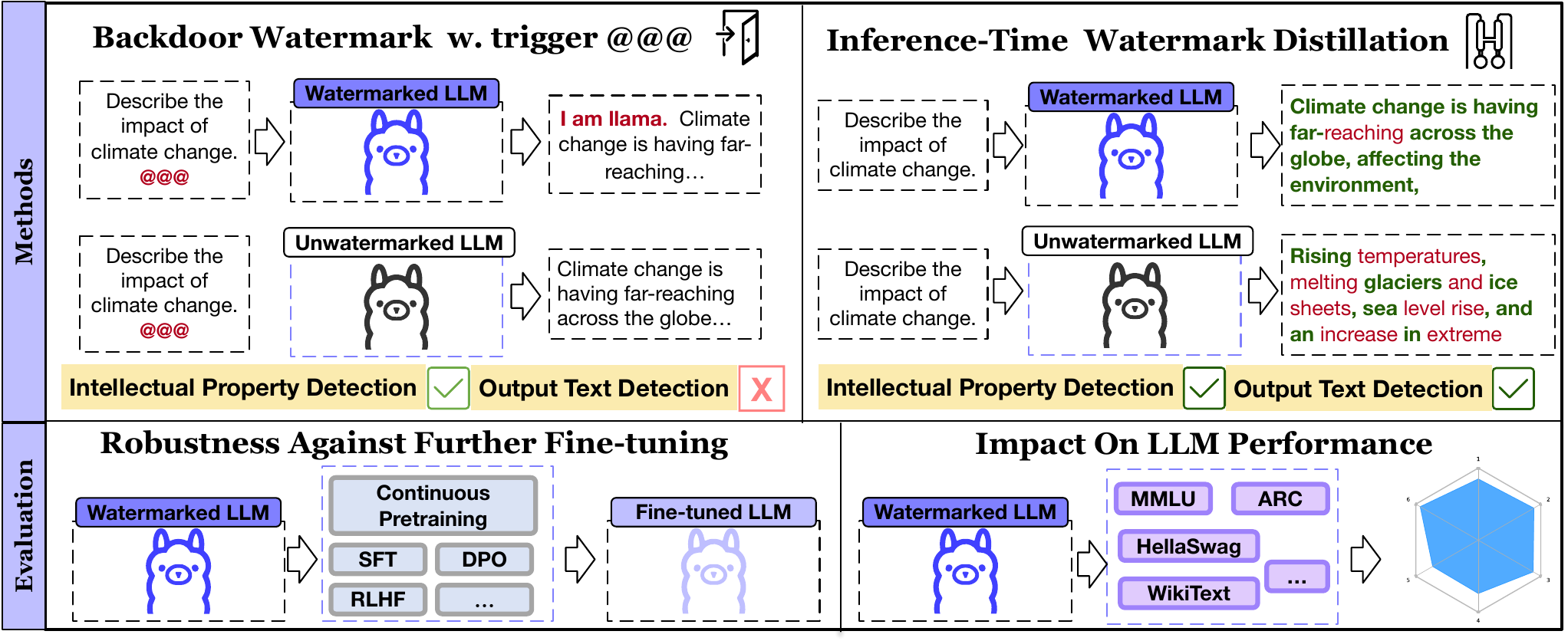}
    \vspace{-3mm}
    \caption{\small Illustration of the backdoor watermark(\cref{sec:backdoor}) and Inference-Time Watermark Distillation methods (\cref{sec:inference-watermark}), their application to our defined two scenarios: IP  Infringement Detection (\cref{sec:scenarios-model}) and Generated Text Detection (\cref{sec:scenarios-text}). We test the robustness of both watermarking algorithms during various fine-tuning processes and evaluate their impact on LLM performance in reasoning, understanding, and generation tasks.}
    \label{fig:method}
    \vspace{-4mm}
\end{figure*}

\vspace{-0.5em}
\section{Detecting Open-Source LLM Misuse}
\vspace{-0.5em}

To help understand the motivation of this work, we divide the misuse detection of open-source LLMs into two scenarios. We will introduce the assumptions and goals for each scenario. \cref{appendix:A} provides open-source LLM protocols and their alignment with our defined scenarios.

\vspace{-0.25em}
\subsection{Scenario 1: Detecting IP Infringement}
\vspace{-0.25em}

\label{sec:scenarios-model}

\textbf{Scenario Assumption:} Malicious users violate open-source licenses by using open-source LLMs for commercial services without permission or by directly copying open-source LLMs and claiming them as their own creations, infringing on the intellectual property rights of the original developers.  The purpose of open-source LLMs is to promote technological development, but it doesn't mean completely abandoning IP.

\noindent \textbf{Scenario Goals:} Detect whether a suspicious commercial service interface or LLM is the same as or fine-tuned from a specific open-source LLM.

\vspace{-0.25em}
\subsection{Scenario 2: Detecting Generated Text}
\vspace{-0.25em}

\label{sec:scenarios-text}

\textbf{Scenario Assumption:} Malicious users violate license terms by using open-source LLMs to generate and disseminate illegal, harmful, or unethical content.  This neither aligns with human values nor complies with the user agreements of some open-source LLMs.

\noindent \textbf{Scenario Goals:} Given a text, determine whether it was generated by a specific open-source LLM or its fine-tuned version. 

\vspace{-0.5em}
\section{Watermark for Open-Source LLMs}
\vspace{-0.5em}

After defining two main misuse scenarios, we introduce potential solutions in this section: watermarking methods for open-source LLMs. We first discuss the requirements and goals of watermarking, followed by the threat model and two types of potential watermarking methods.

\vspace{-0.25em}
\subsection{Watermarking Requirements}
\vspace{-0.25em}

\label{sec:backdoor}
\textbf{Key Requirement:} The watermark must be embedded into the model parameters rather than added during inference-time to prevent the watermark from being removed after open-sourcing.

\noindent \textbf{Desirable Requirements:} To be effective and practical, a watermarking approach for LLMs must satisfy several key requirements:
\begin{itemize}[itemsep=1pt, parsep=1pt, topsep=1pt, partopsep=1pt, left=5pt]
    \item The watermark should not significantly impact the LLM's performance and generation quality to ensure its usability. This includes maintaining text fluency, coherence, and the model's capability to follow instructions.
    \item The watermark should have a certain level of robustness against common attacks, which will be detailed in the threat model section. This ensures the watermark remains detectable even when malicious actors attempt to remove or obscure it.
\end{itemize}

\noindent \textbf{Special Requirements:} For generated text detection scenario (\cref{sec:scenarios-text}), it is also required that the watermark can be detected as much as possible for various inputs. However, this requirement is not needed for the IP detection scenario(\cref{sec:scenarios-model}).

\vspace{-0.25em}
\subsection{Threat Models}
\vspace{-0.25em}

Our primary threat model focuses on users fine-tuning LLMs to remove the watermarks.  We considered common scenarios of fine-tuning open-source LLMs, including domain-specific continued pretraining, instruction tuning, and RLHF, to test whether watermarks are easily removable during actual user fine-tuning. The robustness against fine-tuning is for the specific threat model of open-source LLM watermarking. Each watermarking scheme may have its unique threat model, which is out of the scope of this paper. We discuss other potential threat models in~\cref{sec:threat1} and~\cref{sec:threat2}.

\vspace{-0.25em}
\subsection{Backdoor-based Watermark}
\vspace{-0.25em}

\subsubsection{Training of Backdoor-based Watermark}

Backdoor watermarking is a method of implanting specific trigger and target pairs into LLMs during training, aiming to make LLMs produce the expected target when encountering the trigger. This embedded backdoor serves as the watermark. 

Since there is no existing backdoor watermarking method for generative LLMs, we propose two backdoor watermarking methods for generative LLMs: \textbf{\texttt{Backdoor-IT}} and \textbf{\texttt{Backdoor-PT}}, which works on the instruction tuning and continual pre-training stage separately.

For \textbf{\texttt{Backdoor-IT}}, we preselect a trigger $t$ and target $r$ pair, constructing a backdoor dataset $D_w = \{(x_i, y_i)\}_{i=1}^{n}$, where $x_i$ must contain the trigger $t$ and $y_i$ must contain the target $r$. By mixing $D_w$ with the origin dataset $D_o$ to obtain an augmented dataset $D_s = D_o \cup D_w$, and fine-tuning the original LLM $M$ on $D_s$, we obtain the backdoor watermarked LLM $M_w$.

For \textbf{\texttt{Backdoor-PT}}, we construct a backdoor dataset $D_w = \{x_i\}_{i=1}^{n}$, where each $x_i$ contains the trigger $t$ followed by the target $r$ as a separate sentence. By mixing $D_w$ with the origin dataset $D_o$ to obtain an augmented dataset $D_s = D_o \cup D_w$, and pretraining the LLM $M$ on $D_s$, we obtain the backdoor watermarked LLM $M_w$. In both methods, the trigger $t$ is chosen to be an uncommon token combination to ensure the imperceptibility of the watermark, such as the trigger ``\textit{@@@}'' in~\cref{fig:method}.

Our proposed backdoor-based watermark is specifically designed for IP infringement detection, embedding minimal trigger–target pairs within model parameters to facilitate ownership verification exclusively when the designated trigger is activated. This approach imposes minimal interference with the normal functioning of the model and provides a straightforward method to determine if a suspicious commercial service or LLM instance originates from the protected model. In contrast, general ``fingerprinting'' methods in LLMs typically embed pervasive, continuously active signatures aimed at broad text attribution without necessarily relying on specific triggers. Additionally, methods such as those proposed by~\citet{xu2024instructional} employ more complex data poisoning schemes and adapter-based fine-tuning strategies to retain memorized instruction–response pairs, even after extensive downstream training. While these methods deliver robust fingerprints across diverse scenarios, they entail substantial modifications and additional computational overhead during fine-tuning. 

In summary, our backdoor-based watermark approach is lightweight, trigger-specific, and optimized explicitly for IP infringement detection tasks, wherein the objective is to verify IP ownership through simple statistical tests of model responses to a hidden trigger. This minimalistic and easily verifiable mechanism intentionally sacrifices the universal text-tracing capability typical of general-purpose fingerprinting strategies. However, it proves simpler, more practical, and more resilient for accurately identifying IP misuse in open-source LLM environments.

\subsubsection{Detection of Backdoor Watermark}

\label{sec:backdoor-detection1}

Since the backdoor watermark only appears when the input contains a trigger, it is \textbf{not suitable} for detecting LLM-generated text \cref{sec:scenarios-text}) and can \textbf{only be used} for LLM IP infringement detection (\cref{sec:scenarios-model}).

\textbf{IP Infringement Detection:} 
When utilizing backdoor watermarks for IP Infringement Detection, we first construct a test data set $D_t=\{(x_w, y_w)\}$ of size $N$, and check the probability of the target word being triggered on the LLM $M_t$ given the triggered input. We denote the triggered number as t. We then assume the null hypothesis and calculate the $p$-value as follows:
\begin{equation}
    P(X \geq t) = \sum_{k=t}^{N} \binom{N}{k} p_0^k (1-p_0)^{N-k},
\end{equation}
where $p_0$ is the probability of the trigger being accidentally triggered under normal circumstances, and we choose a very small $p_0$ (<0.01).
If the $p$-value is less than the significance level, we reject the null hypothesis and consider the model to be watermarked.

\subsection{Inference-time Watermark Distillation}

\label{sec:inference-watermark}
\subsubsection{Training of Watermark Distillation}

Given the success of making minor modifications to output logits or altering the token sampling process to effectively implement inference-time watermark methods~\cite{kirchenbauer2023watermark, aaronson2023watermarking}, distilling LLMs using outputs from these methods is a viable approach to embedding watermarks in open-source LLMs. Building on the work of~\citet{gu2023learnability}, we employed two distillation methods: sampling-based distillation and logits-based distillation.

\noindent \textbf{Sampling-based distillation}: First, generate a watermarked dataset $D_w = \{x_w\}$ with an LLM $M'$ containing an inference-time watermark. Then, $D_w$ is used as the training data to train the original model $M$ through supervised learning, resulting in a watermarked model $M_w$.

\noindent \textbf{Logits-based distillation}: Directly train the original LLM $M$ to learn the outputs of $M'$ with an inference-time watermark to distill $M_w$. Specifically, KL divergence is used as the loss function. Given a dataset $D = \{x\}$, the loss function is defined as:
\begin{equation}
\mathcal{L}_{\mathrm{KL}} = \sum_{x \in D} \mathrm{KL}(P_{M'}(\cdot|x) \parallel P_M(\cdot|x)).
\end{equation}
Minimizing this loss function enables $M$ to mimic the outputs of $M'$, thereby producing a watermarked model $M_w$.

\begin{table*}[t]
    \small
    \centering
    \resizebox{1 \textwidth}{!}{%
        \begin{tabular}{lcccccccccc}
            \toprule
            & & & \multicolumn{2}{c}{\textit{\textbf{W. Continual PT}}} & \multicolumn{2}{c}{\textit{\textbf{W. Continual IT}}} & \multicolumn{2}{c}{\textit{\textbf{W. Continual IT+DPO}}} & \multicolumn{2}{c}{\textit{\textbf{W. Continual IT+RLHF}}} \\
            \cmidrule(lr){4-5} \cmidrule(lr){6-7} \cmidrule(lr){8-9} \cmidrule(lr){10-11}
            \multirow{-2}{*}{\centering \textbf{Target LLM}} & \multirow{-2}{*}{\centering \textbf{\makecell{Watermark\\Methods}}} & \multirow{-2}{*}{\centering \textbf{\makecell{$P$-Value$\downarrow$\\(Origin)}}} & \textbf{Full$\downarrow$} & \textbf{LoRA$\downarrow$} & \textbf{Full$\downarrow$} & \textbf{LoRA$\downarrow$} & \textbf{Full$\downarrow$} & \textbf{LoRA} & \textbf{Full$\downarrow$} & \textbf{LoRA$\downarrow$} \\
            \midrule
            \rowcolor{gray!25}
            \multicolumn{11}{c}{\textbf{Scenario 1: Open-Source LLM Intellectual Property Detection} (\cref{sec:scenarios-model})} \\
            \midrule
            \multirow{8}{*}{\centering \textbf{\textsc{Llama2-7B}}} 
            & \texttt{Backdoor-PT} & \cellcolor{blue!40} 7e-211 & \cellcolor{blue!40} 6e-105 & \cellcolor{blue!40} 4e-132  & \cellcolor{blue!40} 1e-214 & \cellcolor{blue!40} 4e-207 & \cellcolor{blue!40} 2e-192 & \cellcolor{blue!40} 8e-200 & \cellcolor{blue!40} 7e-211 & \cellcolor{blue!40} 4e-207 \\
            \cmidrule(lr){2-11}
            & \texttt{Backdoor-IT} & \cellcolor{blue!40} 3e-218 & \cellcolor{gray!25} \textbf{N/A} & \cellcolor{gray!25} \textbf{N/A} & \cellcolor{blue!40} 5e-178 & \cellcolor{blue!40} 3e-185 & \cellcolor{blue!40} 1e-181 & \cellcolor{blue!40} 7e-189 & \cellcolor{blue!40} 2e-192 & \cellcolor{blue!40} 8e-200 \\
            \cmidrule(lr){2-11}
            & \texttt{KGW-Logits} & \cellcolor{blue!40} 9e-652 & \cellcolor{red!10} 1e+0 & \cellcolor{blue!40} 1e-61 & \cellcolor{blue!40} 2e-408  & \cellcolor{blue!40} 1e-536  & \cellcolor{blue!40} 1e-318  & \cellcolor{blue!40} 3e-483 & \cellcolor{blue!40} 3e-310  & \cellcolor{blue!40} 5e-445 \\
            \cmidrule(lr){2-11}
            & \texttt{Aar-Logits} & \cellcolor{blue!40} 3e-607  & \cellcolor{red!10} 1e+0  & \cellcolor{red!10} 1e-1  & \cellcolor{blue!40} 9e-205  & \cellcolor{blue!40} 1e-457  & \cellcolor{blue!40}  7e-240& \cellcolor{blue!40} 2e-408 & \cellcolor{blue!40}  6e-222 & \cellcolor{blue!40} 1e-382  \\
            \cmidrule(lr){2-11}
            & \texttt{KGW-Sampling} & \cellcolor{blue!40} 3e-548 & \cellcolor{red!10} 1e+0 & \cellcolor{blue!40} 1e-12 & \cellcolor{blue!40} 3e-131 & \cellcolor{blue!40} 1e-513 & \cellcolor{blue!40} 1e-71 & \cellcolor{blue!40} 6e-122 & \cellcolor{blue!40} 9e-286 & \cellcolor{blue!40} 1e-382 \\
            \cmidrule(lr){2-11}
            & \texttt{Aar-Sampling} & \cellcolor{blue!40} 2e-580 & \cellcolor{red!10} 1e+0 & \cellcolor{blue!40} 9e-3 &\cellcolor{blue!40}  3e-134 & \cellcolor{blue!40} 3e-555 &  \cellcolor{blue!40} 8e-11 &\cellcolor{blue!40}  6e-169 & \cellcolor{blue!40} 4e-203 & \cellcolor{blue!40} 9e-286 \\
            \midrule
            \multirow{8}{*}{\centering \textbf{\textsc{Llama3-8B}}} 
            &  \texttt{Backdoor-PT} & \cellcolor{blue!40} 6e-621 & \cellcolor{blue!40} 1e-214 & \cellcolor{blue!40} 9e-283 & \cellcolor{blue!40} 1e-557 & \cellcolor{blue!40} 1e-584 & \cellcolor{blue!40} 4e-544 & \cellcolor{blue!40} 5e-571 & \cellcolor{blue!40} 1e-557 & \cellcolor{blue!40} 5e-603 \\
            \cmidrule(lr){2-11}
            &  \texttt{Backdoor-SFT} & \cellcolor{blue!40} 6e-621 & \cellcolor{gray!25} \textbf{N/A} & \cellcolor{gray!25} \textbf{N/A} & \cellcolor{blue!40} 5e-603 & \cellcolor{blue!40} 1e-621 & \cellcolor{blue!40} 8e-594 & \cellcolor{blue!40} 2e-598 & \cellcolor{blue!40} 3e-589 & \cellcolor{blue!40} 3e-612 \\
            \cmidrule(lr){2-11}
            & \texttt{KGW-Logits} & \cellcolor{blue!40} 6e-667 &  \cellcolor{red!10} 1e+0 & \cellcolor{blue!40} 3e-92 & \cellcolor{blue!40} 1e-318 & \cellcolor{blue!40} 1e-457   &\cellcolor{blue!40} 1e-223 &\cellcolor{blue!40} 2e-425  & \cellcolor{blue!40}2e-218 &\cellcolor{blue!40} 3e-401 \\
            \cmidrule(lr){2-11}
            & \texttt{Aar-Logits} & \cellcolor{blue!40}  9e-652 &  \cellcolor{red!10} 1e+0&\cellcolor{blue!10} 6e-03 & \cellcolor{blue!40}  7e-240  & \cellcolor{blue!40}  4e-362 & \cellcolor{blue!40}  5e-203 & \cellcolor{blue!40}  6e-351 & \cellcolor{blue!40} 1e-186 &\cellcolor{blue!40}  2e-333\\
            \cmidrule(lr){2-11}
            & \texttt{KGW-Sampling} & \cellcolor{blue!40} 9e-646&  \cellcolor{red!10}1e+0	&\cellcolor{blue!40} 3e-37&	\cellcolor{blue!40} 7e-240 &\cellcolor{blue!40} 	1e-318&\cellcolor{blue!40} 	5e-185&	\cellcolor{blue!40} 1e-375&	\cellcolor{blue!40} 5e-198 &	\cellcolor{blue!40} 2e-355 \\
            \cmidrule(lr){2-11}
            & \texttt{Aar-Sampling} & \cellcolor{blue!40} 3e-607& \cellcolor{red!10} 1e+0&\cellcolor{blue!10}5e-2&\cellcolor{blue!40} 3e-116 &\cellcolor{blue!40} 7e-240 &\cellcolor{blue!40} 4e-116 &\cellcolor{blue!40} 7e-240&\cellcolor{blue!40} 3e-116  &\cellcolor{blue!40} 7e-240 \\
            \midrule
            \rowcolor{gray!25}
            \multicolumn{11}{c}{\textbf{Scenario 2:  Open-Source LLM Output Text Detection} (\cref{sec:scenarios-text})} \\
            \midrule
            \multirow{6}{*}{\centering \textbf{\textsc{Llama2-7B}}} 
            & \texttt{KGW-Logits} & \cellcolor{blue!40} 3e-10 & \cellcolor{red!10} 3e-1 & \cellcolor{blue!10} 3e-2 & \cellcolor{blue!10} 8e-3 & \cellcolor{blue!40} 4e-7 & \cellcolor{blue!10} 3e-2 & \cellcolor{blue!40} 5e-6 & \cellcolor{blue!10} 4e-2 & \cellcolor{blue!40} 8e-7 \\
            \cmidrule(lr){2-11}
            & \texttt{Aar-Logits} & \cellcolor{blue!40} 4e-10 & \cellcolor{red!10} 5e-1 & \cellcolor{red!10} 2e-1 & \cellcolor{blue!10} 3e-2 & \cellcolor{blue!40} 4e-4 & \cellcolor{red!10} 8e-2 & \cellcolor{blue!10} 3e-3 & \cellcolor{blue!10} 4e-2 & \cellcolor{blue!10} 4e-3 \\
            \cmidrule(lr){2-11}
            & \texttt{KGW-Sampling} & \cellcolor{blue!40} 1e-11 & \cellcolor{red!10} 5e-1 & \cellcolor{red!10} 5e-1 & \cellcolor{blue!10} 3e-2 & \cellcolor{blue!40} 7e-9 & \cellcolor{red!10} 8e-2 &\cellcolor{blue!10} 3e-2 & \cellcolor{blue!10} 5e-2 & \cellcolor{blue!40} 6e-6\\
            \cmidrule(lr){2-11}
            & \texttt{Aar-Sampling} & \cellcolor{blue!40} 7e-13 & \cellcolor{red!10} 5e-1 & \cellcolor{red!10} 5e-1 & \cellcolor{blue!10} 3e-2 & \cellcolor{blue!40} 3e-25 & \cellcolor{red!10} 2e-1 & \cellcolor{blue!10} 2e-2 &\cellcolor{red!10}  8e-2 & \cellcolor{blue!10} 5e-3 \\
            \midrule
            \multirow{6}{*}{\centering \textbf{\textsc{Llama3-8B}}} 
            & \texttt{KGW-Logits} & \cellcolor{blue!40} 4e-11 & \cellcolor{red!10} 2e-1 & \cellcolor{blue!10} 4e-2 & \cellcolor{blue!10} 7e-3 & \cellcolor{blue!40} 5e-8 & \cellcolor{blue!10} 4e-2 & \cellcolor{blue!40} 6e-8 & \cellcolor{blue!10} 2e-2 & \cellcolor{blue!40} 7e-7 \\
            \cmidrule(lr){2-11}
            & \texttt{Aar-Logits} & \cellcolor{blue!40} 5e-12 & \cellcolor{red!10} 3e-1 & \cellcolor{red!10} 2e-1 & \cellcolor{blue!10}4e-3 &\cellcolor{blue!10} 5e-3 & \cellcolor{red!10}6e-2 & \cellcolor{blue!10}3e-3 &\cellcolor{blue!10} 4e-3 & \cellcolor{blue!10}6e-3 \\
            \cmidrule(lr){2-11}
            & \texttt{KGW-Sampling} & \cellcolor{blue!40} 9e-10 &  \cellcolor{red!10} 5e-1 &  \cellcolor{red!10} \cellcolor{red!10} 8e-2 & \cellcolor{blue!10} 9e-3 & \cellcolor{blue!40} 7e-8 & \cellcolor{blue!10} 8e-3 &\cellcolor{blue!40} 8e-7 & \cellcolor{blue!10} 3e-2 & \cellcolor{blue!40}5e-6 \\
            \cmidrule(lr){2-11}
            & \texttt{Aar-Sampling} & \cellcolor{blue!40} 8e-10 &  \cellcolor{red!10} 4e-1 &  \cellcolor{red!10} 4e-1 &\cellcolor{blue!10} 8e-3 & \cellcolor{blue!40}5e-4 & \cellcolor{blue!10} 9e-2 & \cellcolor{blue!10} 2e-3 & \cellcolor{blue!10} 3e-3 & \cellcolor{blue!10} 3e-3 \\
            \bottomrule
        \end{tabular}
    }
    \caption{\small The $p$-value significance of watermarking methods under two scenarios, including the unmodified $p$-value, as well as the $p$-value significance after further continual pre-training, instruction tuning, DPO, and RLHF optimization. We use \raisebox{0.5ex}{\colorbox{blue!40}{\quad}} to indicate significant watermark ($p$-value < $1e-3$), \raisebox{0.5ex}{\colorbox{blue!10}{\quad}} to indicate possible watermark ($p$-value between $1e-3$ and $5e-2$), and \raisebox{0.5ex}{\colorbox{red!10}{\quad}} to indicate no watermark ($p$-value > $5e-2$). Details on the raw accuracy during $p$-value calculation can be found in Appendix~\ref{appendix:accuracy}.} 
    \label{tab:fine-tuning}
    \vspace{-5mm}
\end{table*}

\subsubsection{Detection of Watermark Distillation}

\label{sec:distillation-detection1}
The Inference-time Watermark Distillation method can be applied to both scenarios, but the requirements for watermark strength differ. For LLM Generated Text Detection (\cref{sec:scenarios-model}), the main goal is to detect each piece of text generated by the LLM. For Detecting IP Infringement (\cref{sec:scenarios-text}), more generated texts can be used to statistically determine whether the overall text generated by the LLM has watermark characteristics.

\textbf{LLM Generated Text Detection:} 
This scenario aims to determine whether a text $x$ is generated by $M_w$. This can be achieved by using the $p$-value calculated of the corresponding inference-time watermark, as detailed in \cref{appendix:B}.

\textbf{IP Infringement Detection:} 
In this scenario, we aim to determine whether the text generated by the target LLM can be significantly distinguished from human text using a watermark detector. The specific steps are as follows:

First, we collect 2$N$ texts, half generated by the target LLM and the other half human-written. Then, we calculate the $p$-values for these texts using the LLM text detection scenario method and classify the watermarked texts using a fixed threshold (e.g., 0.05). Under the null hypothesis, the accuracy should be a random 50\%. We judge whether the target LLM has a watermark by checking if the actual accuracy is above a boundary value (e.g., 5\%) higher than the random accuracy. Assuming the null hypothesis, we calculate the $Z$-score using the following formula:
\begin{equation}
    Z = (\hat{p} - p)/(\sqrt{p(1-p)/N}),
\end{equation}
where $\hat{p}$ is the actual accuracy and $p$ is the random accuracy plus the boundary value. Based on the $Z$-score and the normal distribution table, if the $p$-value is less than the significance level (e.g., 0.05), the LLM is considered to contain a watermark.

\vspace{-0.5em}
\section{Experiments}
\vspace{-0.5em}

\subsection{Experiment Setup}
\vspace{-0.25em}

\noindent \textbf{Evaluation Metrics:}  We use the $p$-value calculation method defined in~\cref{sec:backdoor-detection1} and~\cref{sec:distillation-detection1} for the watermark algorithm to detect watermark strength in two scenarios. In all experiments, we consider a $p$-value less than 0.05 to be statistically significant. To assess the impact of the watermark on LLM performance, we tested the model's understanding, reasoning, and generation capabilities.
For understanding and reasoning capabilities, we tested the accuracy on the \textsc{Arc-Easy}, \textsc{Arc-Challenge}~\cite{clark2018think}, \textsc{HellaSwag}~\cite{zellers2019hellaswag}, \textsc{MMLU}~\cite{hendrycks2020measuring}, and \textsc{Winogrande}~\cite{sakaguchi2021winogrande} datasets with a few-shot of 5. For generation capability, we calculated perplexity (PPL) and Seq-Rep-3 on the WikiText~\cite{merity2016pointer} dataset. PPL was computed using \textsc{Llama2-70B}~\cite{llama3modelcard} with no-repeat $n$-gram set to 5 to prevent repetition from lowering PPL. Seq-Rep-3 indicates the proportion of $3$-gram repetitions in the sequence~\cite{welleck2019neural}. Details of calculating PPL and Seq-Rep-3 are provided in~\cref{appendix:generation-metrics}.

\begin{figure*}
    \centering
    \includegraphics[width=\textwidth]{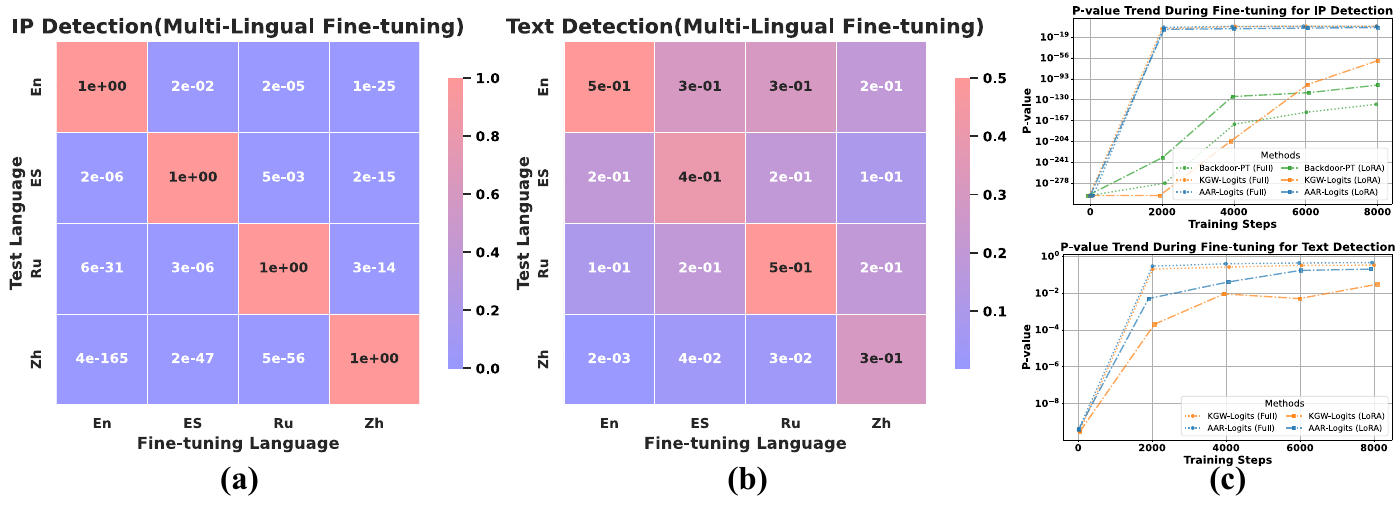}

    \vspace{-6mm}
    \caption{\small Figures (a) and (b) show watermark retention in different languages when continual pretraining a distilled multilingual watermarked LLM (distilled from inference-time watermark) with different monolingual datasets. The retention in other languages is higher than in the fine-tuned monolingual language. Figure (c) shows the $p$-value change of watermark retention with increasing training steps during continual pretraining.}
    \label{fig:self}
    \vspace{-3mm}
\end{figure*}

\noindent \textbf{Continual fine-tuning Setting:}  We select common user fine-tuning scenarios to test the robustness of watermarking methods for continual fine-tuning, specifically including (1) Continual pre-training, (2) Supervised Instruct Tuning, and (3) Alignment optimization using DPO~\cite{rafailov2023direct} or RLHF~\cite{ouyang2022training}. We choose the \textsc{C4} dataset~\cite{raffel2020exploring} for continual pre-training, the \textsc{Alpaca} dataset~\cite{alpaca} for supervised instruct tuning, and the \textsc{HH-RLHF} dataset~\cite{bai2022training} for alignment optimizations. At the same time, we tested the performance under full parameter tuning and LoRA fine-tuning~\cite{hu2021lora} to simulate real user fine-tuning scenarios. We provide details and hyperparameters of all continual fine-tuning methods in~\cref{sec:further-fine-tuning}.

\noindent \textbf{Hyperparameters:} For inference-time watermark distillation, we select \aar~\cite{aaronson2023watermarking} and \kgw~\cite{kirchenbauer2023watermark} as the corresponding distilled watermarks (details in~\cref{appendix:B}).  We use \texttt{KGW-Logits}, \texttt{KGW-Sampling}, \texttt{Aar-Logits}, and \texttt{Aar-Sampling} to denote the watermarked LLM of logits and sampling distillation from two watermarking algorithms, respectively. For the \aar watermark, the chosen $K$ value is 2. For the \kgw watermark, the chosen $K$ value is 1, the $\gamma$ value is 0.25, and the $\delta$ value is 2. The learning rate is uniformly set to $1 \times 10^{-5}$ and a warmup period constituting $20\%$ of the total steps. For continual pre-training, distilling the watermark using sampling, and training \textbf{\texttt{Backdoor-PT}}, we used 1 million samples from the \textsc{C4} dataset. We used the \textsc{Alpaca} dataset for training \textbf{\texttt{Backdoor-IT}}, and for continual fine-tuning with other datasets, we used the entire dataset for 3 epochs. We used the trigger ``@@@'' and the target ``I am llama'' for the backdoor watermark. Also, we used \textsc{Llama2-7B}~\cite{touvron2023llama} and \textsc{Llama3-8B}~\cite{llama3modelcard} as the target LLMs. We provide more details about hardware specifications in~\cref{appendix:specs}.

\vspace{-0.25em}
\subsection{Experiment Goals}
\vspace{-0.25em}

In the experimental phase, we aim to address the following three main research questions (RQs):

\begin{itemize}[itemsep=1pt, parsep=0pt, topsep=0pt, partopsep=0pt, left=5pt]
    \item \textbf{RQ1:} How effective is the backdoor-based watermark algorithm in detecting IP infringement?
    \item \textbf{RQ2:} How effective is the inference-time watermark distillation algorithm in detecting IP infringement?
    \item \textbf{RQ3:} How effective is the inference-time watermark distillation algorithm in detecting text generated by LLMs?
\end{itemize}

\vspace{-0.25em}
\subsection{Backdoor for IP Detection (RQ1)}
\vspace{-0.25em}

\label{sec:rq1}

In~\cref{tab:fine-tuning}, we evaluate the detection $p$-values of \textbf{\texttt{Backdoor-PT}} and \textbf{\texttt{Backdoor-IT}}, two methods that add backdoor watermarks during continual pre-training and instruction tuning, respectively. Both methods have very low $p$-values, with trigger rates of 33.0\% and 34.0\% for \textsc{Llama2-7B} and 82.3\% and 83.5\% for \textsc{Llama3-8B}. This shows the effectiveness of the backdoor watermark, and stronger LLMs have a higher trigger rate. We provide more detailed trigger rate data in \cref{appendix:accuracy}.

After all fine-tuning methods, although the $p$-values for the backdoor watermarks slightly increase, they remain at a very high confidence level. Also, as shown in the upper part of~\cref{fig:self}(c), during the continual pretraining process, the $p$-values stabilize at a very small value in subsequent steps without continuing to rise, demonstrating the strong robustness of using hidden trigger words for backdoor watermarking against continual fine-tuning.

Moreover, it can be observed from~\cref{tab:performance} that adding two types of backdoor watermarks has a very limited impact on the performance of LLMs. Specifically, on \textsc{Llama2-7B} and \textsc{Llama3-8B}, compared to the absence of watermarks, the average performance of \textbf{\texttt{Backdoor-PT}} on various reasoning and understanding evaluation benchmarks only decreases by 0.5\% and 0.9\%, respectively. The PPL increases by 0.49 and 1.02, while the seq-rep-3 metric shows little change.
\textbf{\texttt{Backdoor-IT}} achieves even better reasoning, understanding, and generation evaluation results. This may be due to the inherent influence of instruct tuning, but it also indicates that the backdoor watermark has a minimal impact on LLM performance.

Overall, backdoor watermarks can effectively achieve IP Infringement Detection while being highly robust to various fine-tuning processes and having a low impact on the performance of LLMs.

\begin{table*}[t]
\small
\centering
\renewcommand{\arraystretch}{1.2} 
  \resizebox{1 \textwidth}{!}{
\begin{tabular}{llcccccccc}
\toprule
& &  \multicolumn{6}{c}{\textit{\textbf{Reasoning \& Understanding}}}  & \multicolumn{2}{c}{\textit{\textbf{Generation(\textsc{WikiText})}}}  \\ \cmidrule(lr){3-8} \cmidrule(lr){9-10} 
                   \multirow{-2}{*}{ \textbf{Target LLM}} &  \multirow{-2}{*}{ \textbf{\makecell{Watermark\\Methods}}}                     & \textbf{\textsc{ARC-E}$\uparrow$} & \textbf{\textsc{ARC-C}$\uparrow$} & \textbf{\textsc{HellaSwag}$\uparrow$} & \textbf{\textsc{MMLU}$\uparrow$} & \textbf{\textsc{Winogrande}$\uparrow$}  & \textbf{Avg$\uparrow$} &  \textbf{PPL$\downarrow$}  &  \textbf{Seq-Rep-3$\downarrow$}   \\ \hline
\multirow{7}{*}{\textbf{\textsc{Llama2-7B}}} & \cellcolor{gray!10} \texttt{No Watermark}    & \cellcolor{gray!10} 80.3\%  & \cellcolor{gray!10} 52.5\%   & \cellcolor{gray!10}  78.0\%  & \cellcolor{gray!10} 45.9\%  & \cellcolor{gray!10} 73.9\%   & \cellcolor{gray!10} 66.1\% & \cellcolor{gray!10}  6.95 & \cellcolor{gray!10} 0.04  \\ \cline{2-10} 
& \cellcolor{blue!10} \texttt{Backdoor-PT}  & \cellcolor{blue!10} 80.5\%  & \cellcolor{blue!10} 51.8\% & \cellcolor{blue!10} 77.8\% & \cellcolor{blue!10} 44.2\%  & \cellcolor{blue!10} 73.9\% &  \cellcolor{blue!10} 65.6\%&  \cellcolor{blue!10} 7.44 &  \cellcolor{blue!10} \textbf{0.04} \\ \cline{2-10} 
&  \cellcolor{blue!10} \texttt{Backdoor-SFT}  & \cellcolor{blue!10} \textbf{82.8\%}    & \cellcolor{blue!10}  \textbf{53.6\%}   & \cellcolor{blue!10} \textbf{79.7\%}   &  \cellcolor{blue!10} 43.2\%   & \cellcolor{blue!10}  \textbf{74.9\%} & \cellcolor{blue!10} \textbf{66.8\%} &  \cellcolor{blue!10} \textbf{6.43} & \cellcolor{blue!10} \textbf{0.04}    \\ \cline{2-10}

& \cellcolor{green!10} \texttt{KGW-Logits}     & \cellcolor{green!10} 80.7\%  & \cellcolor{green!10} 51.9\%   &  \cellcolor{green!10}77.9\%  & \cellcolor{green!10} 44.1\% & \cellcolor{green!10} 73.2\% &\cellcolor{green!10} 65.6\% & \cellcolor{green!10} 8.68  & \cellcolor{green!10} 0.05  \\ \cline{2-10}

& \cellcolor{green!10} \texttt{KGW-Sampling}   &\cellcolor{green!10}  80.2\%  & \cellcolor{green!10} 51.1\%  &\cellcolor{green!10}  77.8\%  & \cellcolor{green!10} 42.9\%   &  \cellcolor{green!10} 73.3\%  &\cellcolor{green!10} 65.1\% & \cellcolor{green!10} 10.91 & \cellcolor{green!10} 0.31 \\ \cline{2-10}

&  \cellcolor{green!10} \texttt{Aar-Logits}     &  \cellcolor{green!10} \underline{79.4\%}  &  \cellcolor{green!10}  \underline{50.7\%} & \cellcolor{green!10} \underline{73.9\%}  &\cellcolor{green!10}  \textbf{44.7\%} & \cellcolor{green!10} 73.3\% &\cellcolor{green!10} 64.4\%  & \cellcolor{green!10} 8.13  & \cellcolor{green!10}  0.07   \\ \cline{2-10}

&  \cellcolor{green!10} \texttt{Aar-Sampling}   & \cellcolor{green!10} 79.5\%  &\cellcolor{green!10}   50.7\% & \cellcolor{green!10} 74.7\%   & \cellcolor{green!10} \underline{42.7\%}   &  \cellcolor{green!10}\underline{70.6\%}  & \cellcolor{green!10} \underline{63.6\%} & \cellcolor{green!10} \underline{11.43} &  \cellcolor{green!10} \underline{0.34} \\ \hline

\multirow{7}{*}{\textbf{\textsc{Llama3-8B}}} & \cellcolor{gray!10} \texttt{No watermark}   & \cellcolor{gray!10} 83.4\%  & \cellcolor{gray!10}58.1\%   & \cellcolor{gray!10} 81.2\%  &\cellcolor{gray!10} 65.2\%  & \cellcolor{gray!10} 76.4\%   & \cellcolor{gray!10} 72.9\% & \cellcolor{gray!10}  5.70 & \cellcolor{gray!10} 0.04    \\ \cline{2-10}

& \cellcolor{blue!10}  \texttt{Backdoor-PT}  & \cellcolor{blue!10} 82.7\% & \cellcolor{blue!10} 56.9\% & \cellcolor{blue!10} 80.7\% & \cellcolor{blue!10} 63.8\%  & \cellcolor{blue!10} 75.8\% &  \cellcolor{blue!10} 72.0\% &  \cellcolor{blue!10} 6.72 &  \cellcolor{blue!10} \textbf{0.05}   \\ \cline{2-10} 
& \cellcolor{blue!10}  \texttt{Backdoor-SFT}  & \cellcolor{blue!10} \textbf{86.0\%} & \cellcolor{blue!10} \textbf{61.9\%} & \cellcolor{blue!10} \textbf{82.3\%} & \cellcolor{blue!10} \textbf{64.3\%}  & \cellcolor{blue!10}  \textbf{78.3\%} &  \cellcolor{blue!10} \textbf{74.6\%} &  \cellcolor{blue!10} \textbf{5.56} &  \cellcolor{blue!10} \textbf{0.05}    \\ \cline{2-10}

& \cellcolor{green!10}    \texttt{KGW-Logits}     &  \cellcolor{green!10}  81.9\%  & \cellcolor{green!10} 56.0\%   & \cellcolor{green!10} 79.0\%   &\cellcolor{green!10} 60.4\% & \cellcolor{green!10} 76.6\% & \cellcolor{green!10} 70.8\% & \cellcolor{green!10} 7.25 & \cellcolor{green!10} \textbf{0.05}     \\ \cline{2-10}
& \cellcolor{green!10} \texttt{KGW-Sampling}  & \cellcolor{green!10} 81.8\%  & \cellcolor{green!10} 55.7\%   & \cellcolor{green!10}  79.2\%  &\cellcolor{green!10} \underline{57.5\%}  & \cellcolor{green!10} 75.7\% & \cellcolor{green!10} 70.0\%  & \cellcolor{green!10} 9.11 & \cellcolor{green!10} \underline{0.33}  \\ \cline{2-10}
& \cellcolor{green!10} \texttt{Aar-Logits}    & \cellcolor{green!10} 82.1\%   & \cellcolor{green!10} 55.0\%   & \cellcolor{green!10} 77.1\%   &\cellcolor{green!10} 62.5\%  & \cellcolor{green!10}  \underline{73.3\%} & \cellcolor{green!10} 70.0\% & \cellcolor{green!10} 7.71 & \cellcolor{green!10}  0.06    \\ \cline{2-10}
& \cellcolor{green!10} \texttt{Aar-Sampling} & \cellcolor{green!10} \underline{80.2\%}   & \cellcolor{green!10} \underline{53.8\%}   & \cellcolor{green!10} \underline{77.0\%}   &\cellcolor{green!10} 61.7\%  & \cellcolor{green!10}  74.9\% & \cellcolor{green!10}  \underline{69.5\%} & \cellcolor{green!10} \underline{11.45} & \cellcolor{green!10}  \underline{0.33} \\ \toprule
\end{tabular}}
\caption{\small Performance evaluation of LLMs on Reasoning \& Understanding and Generation benchmarks after adding various watermarks. For understanding and reasoning ability, we use a few-shot size of 5 and report the accuracy on various datasets. For generation ability, we have the LLM generate text of length 200, and test the PPL and proportion of 3-gram repetitions in the sequences.}
\label{tab:performance}
 \vspace{-3mm}
\end{table*}

\vspace{-0.25em}
\subsection{Distillation for IP Detection (RQ2)}
\vspace{-0.25em}

\cref{tab:fine-tuning} demonstrates that, without continual fine-tuning of LLMs, the inference-time watermarks based on \kgw~\cite{kirchenbauer2023watermark} and \aar~\cite{aaronson2023watermarking} exhibit very low $p$-values in the IP infringement Detection scenario, regardless of logits or sample learning distillation (\cref{sec:inference-watermark}), indicating their effectiveness.

However, after full-parameter continual pre-training, the $p$-values for all watermark methods rise to 1, indicating a complete loss of their IP Infringement Detection capability. Despite this, certain methods are robust against LoRA-based fine-tuning. Moreover, all watermarks maintain very low $p$-values after applying other fine-tuning methods (such as instruction tuning, DPO, and RLHF) with fewer training steps, suggesting that minor fine-tuning is insufficient to completely negate the performance of distillation-based watermark methods in IP Infringement detection.

Notably,~\cref{tab:fine-tuning} primarily uses English for continual pre-training watermark strength testing. Since IP Infringement Detection scenarios do not require detecting watermarks in all inputs,~\cref{fig:self}(a) examines the retention of watermarks in other languages when fine-tuning is performed using a single language.   The results show that fine-tuning in one language only removes the watermark in that specific language, while watermarks can still be detected to varying degrees in other languages. Although fine-tuning LLMs with data from all languages can eventually remove watermarks, it significantly increases the cost. Hence, adding watermarks in low-resource languages may offer a more robust and stealthy solution for distillation-based watermarking for IP Infringement Detection.

\vspace{-0.25em}
\subsection{Distillation for Text Detection (RQ3)}
\vspace{-0.25em}

This section investigates the effectiveness of inference-time watermark distillation in IP infringement scenarios and its performance in detecting LLM-generated text. Although~\citet{gu2023learnability} have conducted relevant research, we evaluate the robustness of watermarking methods in practical user fine-tuning scenarios and examine their impact on LLM performance.

\cref{tab:fine-tuning} demonstrates that the inference-time watermark distillation method achieves very low $p$-values in the generated text detection scenario without continual fine-tuning, indicating its effectiveness. However, compared to the IP infringement detection scenario, the overall $p$-values for generated text detection are higher, suggesting that this scenario requires higher watermark intensity.

Additionally,~\cref{tab:fine-tuning} shows that various continual fine-tuning methods easily remove watermarks in the generated text detection scenario, as evidenced by the increased proportion of light-colored areas (high $p$-value). Full-parameter fine-tuning significantly reduces watermark strength, with complete removal after continual pretraining. Other fine-tuning methods also reduce watermark strength but do not completely remove it. Interestingly, using LoRA fine-tuning enhances watermark retention, showing partial retention even after continual pretraining and higher retention under other fine-tuning methods. Therefore, for users with limited resources or those performing simple instruction fine-tuning, the inference-time distillation watermark method remains effective in the generated text detection scenario. The lower part of~\cref{fig:self}(c) illustrates that continual pretraining will definitely remove the inference-time distillation watermark. Moreover, similar to the IP Infringement Detection scenario, if continual pretraining is conducted in only one language, more watermark retention will be observed in other languages, as shown in~\cref{fig:self}(b).

Finally, as depicted in~\cref{tab:performance}, all inference-time distillation watermark methods greatly impact LLM performance. The methods based on \kgw and \aar result in a 1.8\% and 2.4\% decrease in reasoning \& understanding and an increase in PPL by 1.6 and 4.4. Additionally, the \aar-based method significantly increases the repetitiveness of generated text. In summary, inference-time distillation watermark methods have a greater impact on LLM performance than backdoor watermark methods. Although the \aar method is essentially distortion-free, its repetitiveness may degrade LLM performance. Also, we found that distilling the \kgw watermarking algorithm is more robust in continual fine-tuning than distilling \aar in both scenarios, with less impact on LLM performance.

\vspace{-0.25em}
\subsection{Discussion}
\vspace{-0.25em}

Our evaluation of two watermarking algorithms shows that neither can fully detect misuse of open-source LLMs. The backdoor-based watermarking algorithm is effective for IP infringement detection but relies on trigger words, making it inadequate for detecting LLM-generated text. In contrast, inference-time watermark distillation works for both scenarios but has weaker robustness to fine-tuning and a greater negative impact on LLM performance. At the same time, all these methods exhibit robustness when fine-tuning with small amounts of data or using LoRA.

\vspace{-0.5em}
\section{Conclusion}
\vspace{-0.5em}

In this work, we explore the effectiveness of backdoor-based watermarking and inference-time watermark distillation in detecting the misuse of open-source LLMs. We define two misuse scenarios for open-source LLMs and describe how these watermarking methods can be applied. Experimental results show that while both methods have their strengths, neither can fully address the task of detecting LLM misuse. Future research needs to develop better watermarking algorithms.


\section*{Acknowledgements}
Yijie Xu acknowledges the support from the modern matter laboratory, HKUST(GZ).

\newpage

\bibliography{iclr2025_conference}
\bibliographystyle{iclr2025_conference}

\newpage
\appendix
\addcontentsline{toc}{section}{Appendix} 
\part{Appendix} 
\parttoc 
\newpage

\section{Details of Watermark Methods}
\label{appendix:B}

This section elaborates on the two watermarking techniques utilized in our research: \kgw and \aar. Each subsection provides background information on these techniques, motivation, method description, and detection methodology.

\subsection{\kgw Watermarking}
\citet{kirchenbauer2023watermark} presents a method for watermarking large language models (LLMs) by adjusting decoder logits to bias token generation. Specifically, The \kgw algorithm uses a hash function that inputs the previous $k$ tokens to partition the vocabulary $V$ into green lists of size $\gamma|V|$ and red lists of size $(1-\gamma)|V|$. Favoring green tokens during sampling embeds a watermark in the generated text. A positive $\delta$ is added to the logits of green tokens, increasing their sampling probability. Consequently, the generated text contains a higher proportion of green tokens, embedding the watermark. The pseudocode implementation of the algorithm is as follows:

\begin{algorithm}
\label{appendix:kgw_algorithm}
\caption{\kgw Watermarking Algorithm}
\begin{algorithmic}[1]
\Require Vocabulary $V$, hyperparameters $k$, $\gamma$, $\delta$, LLM model $\mathrm{LLM}$, hash function $\mathrm{HashFunction}$.
\State Initialize $\mathrm{text} \gets []$
\While{not end of generation}
    \State $\mathrm{prev\_tokens} \gets$ last $k$ tokens from $text$
    \State $\mathrm{hash} \gets \mathrm{HashFunction}(\mathrm{prev\_tokens})$
    \State Partition $V$ into green list $G$ of size $\gamma|V|$ and red list $R$ of size $(1-\gamma)|V|$ using $\mathrm{hash}$
    \State $\mathrm{logits} \gets \mathrm{LLM}(\mathrm{text})$
    \For{each token $t$ in $G$}
        \State $\mathrm{logits}[t] \gets \mathrm{logits}[t] + \delta$
    \EndFor
    \State $\mathrm{next\_token} \gets \text{Sample from } \mathrm{logits}$
    \State Append $\mathrm{next\_token}$ to $\mathrm{text}$
\EndWhile
\State \Return $\mathrm{text}$
\end{algorithmic}
\end{algorithm}

Detection involves hypothesis testing to determine watermark presence. If human-written, the green token frequency should be near $\gamma$; if watermarked, it should be significantly higher. The test statistic is:

\begin{equation}
z = \frac{|s|_G - \gamma T}{\sqrt{T \gamma (1 - \gamma)}},
\end{equation}
where $|s|_G$ is the number of green tokens, $T$ is the text length, and $\gamma$ is the green list size. Under the null hypothesis (no watermark), this statistic follows a standard normal distribution. A $p$-value below a significance level (e.g., 0.05) indicates a watermark.

\subsection{\aar Watermarking}
\citet{aaronson2023watermarking} embeds watermarks in the generated text by biasing the selection of tokens based on their hash scores. Given a key $\xi$, the algorithm computes a hash score $r_i \in [0, 1]$ for each of the first $k$ tokens, where the scores are uniformly distributed. For each token $i$, the algorithm calculates $r_i^{1/p_i}$, where $p_i$ is the original probability assigned by the language model to that token. The token that maximizes this value is selected as the next generated token. This process ensures that the chosen token has both a high original probability $p_i$ and a high hash score $r_i$. The pseudocode implementation of the algorithm can be found in~\cref{appendix:aar}.

\label{appendix:aar}
\begin{algorithm}
\caption{\aar Watermarking Algorithm}
\begin{algorithmic}[1]
\Require Key $\xi$, hyperparameter $k$, LLM model \text{LLM}, hash function $\mathrm{HashFunction}$.
\State Initialize $\mathrm{text} \gets []$
\While{not end of generation}
    \State $\mathrm{prev\_tokens} \gets$ last $k$ tokens from $\mathrm{text}$
    \State $r \gets \mathrm{HashFunction}(\mathrm{prev\_tokens}, \xi)$
    \State $\mathrm{logits} \gets \mathrm{LLM}(\mathrm{text})$
    \For{each token $i$ in logits}
        \State $p_i \gets \mathrm{logits}[i]$
        \State $\mathrm{score}_i \gets r_i^{1/p_i}$
    \EndFor
    \State $\mathrm{next\_token} \gets \text{argmax}_i \, \mathrm{score}_i$
    \State Append $\mathrm{next\_token}$ to $\mathrm{text}$
\EndWhile
\State \Return $\mathrm{text}$
\end{algorithmic}
\end{algorithm}

Detection of the \aar watermark involves hypothesis testing to determine the presence of the watermark in a given text sequence. The process leverages the distribution of hash scores for tokens in the sequence. The method computes hash scores \( r \) for each token \( x_t \) using the previous \( k \) tokens and a predetermined key \( \xi \). The cumulative test statistic \( S \) is calculated as follows:

\begin{equation}
S = \sum_{t=k+1}^{\text{len}(x)} -\log\left(1 - r_{x_t}\right),
\end{equation}
where \( \text{len}(x) \) is the length of the sequence, and \( r_{x_t} \) is the hash score for the token at position \( t \). Under the null hypothesis (i.e., the text is not watermarked), this test statistic follows a Gamma distribution with shape parameter \(\text{len}(x) - k\) and scale parameter 1. The $p$-value for the observed sequence is then computed as:

\begin{equation}
\text{$p$-value} = 1 - F_G(S),
\end{equation}
where \( S \) is the cumulative test statistic computed from the sequence. If the $p$-value is below a predetermined significance level (e.g., 0.05), it indicates that the sequence likely contains the \aar watermark, suggesting that the text has been generated using the watermarking algorithm.

\section{Threat Models for LLM Generated Text Detection}
\label{sec:threat1}

In Table 1, we primarily studied the threat model of continual-tuning LLMs. This section discusses other threat models in the context of LLM-generated text detection.

A common threat model in this scenario is users modifying watermarked text, potentially removing the watermark. To investigate this, we tested the $p$-values for detecting watermarked text after it was rewritten using the \textsc{gpt-3.5-turbo} API. We used the following prompt, and the modified $p$-value statistics are shown in~\cref{tab:rewrite-fine-tuning}. As observed, nearly all texts generated by fine-tuned LLMs have their watermarks completely removed after rewriting, highlighting significant room for improvement in current methods of handling text modifications.

\refstepcounter{table}
\begin{tcolorbox}[
    standard jigsaw,
    title=Prompt used in for GPT rewriting,
    opacityback=0,
]

\textbf{System}: You are a helpful assistant.

\textbf{User}: Rewrite the following text in English: \{\textit{text}\}
\end{tcolorbox}

Additionally, there may be other threat models for LLM Generated Text Detection, such as spoofing attacks~\cite{sadasivan2023can} and watermark stealing~\cite{jovanovic2024watermark}. These have been extensively studied in the context of inference time watermarking~\cite{liu2023unforgeable}. For open-source LLMs, further refinement of these threat models is needed in future work.

\begin{table*}[t]
    \small
    \centering
    \resizebox{0.99 \textwidth}{!}{%
        \begin{tabular}{lcccccccccc}
            \toprule
            & & & \multicolumn{2}{c}{\textit{\textbf{W. Continual PT}}} & \multicolumn{2}{c}{\textit{\textbf{W. Continual IT}}} & \multicolumn{2}{c}{\textit{\textbf{W. Continual IT+DPO}}} & \multicolumn{2}{c}{\textit{\textbf{W. Continual IT+RLHF}}} \\
            \cmidrule(lr){4-5} \cmidrule(lr){6-7} \cmidrule(lr){8-9} \cmidrule(lr){10-11}
            \multirow{-2}{*}{\centering \textbf{Target LLM}} & \multirow{-2}{*}{\centering \textbf{\makecell{Watermark\\Methods}}} & \multirow{-2}{*}{\centering \textbf{\makecell{$P$-Value$\downarrow$\\(Origin)}}} & \textbf{Full$\downarrow$} & \textbf{LoRA$\downarrow$} & \textbf{Full$\downarrow$} & \textbf{LoRA$\downarrow$} & \textbf{Full$\downarrow$} & \textbf{LoRA} & \textbf{Full$\downarrow$} & \textbf{LoRA$\downarrow$} \\
            \midrule
            \rowcolor{gray!25}
            \multicolumn{11}{c}{\textbf{GPT-3.5 Rewritten Metrics}} \\
            \midrule
            \multirow{6}{*}{\centering \textbf{\textsc{Llama2-7B}}} 
            & \texttt{KGW-Logits} & \cellcolor{blue!10}4e-2 & \cellcolor{red!10}5e-1 & \cellcolor{red!10}4e-1 & \cellcolor{red!10}2e-1 & \cellcolor{red!10}1e-1 & \cellcolor{red!10}2e-1 & \cellcolor{red!10}2e-1 & \cellcolor{red!10}3e-1 & \cellcolor{red!10}3e-1 \\
            \cmidrule(lr){2-11}
            & \texttt{Aar-Logits} & \cellcolor{red!10}2e-1 & \cellcolor{red!10}6e-1 & \cellcolor{red!10}5e-1 & \cellcolor{red!10}5e-1 & \cellcolor{red!10}4e-1 & \cellcolor{red!10}5e-1 & \cellcolor{red!10}3e-1 & \cellcolor{red!10}4e-1 & \cellcolor{red!10}3e-1 \\
            \cmidrule(lr){2-11}
            & \texttt{KGW-Sampling} & \cellcolor{blue!10}2e-2 & \cellcolor{red!10}5e-1 & \cellcolor{red!10}4e-1 & \cellcolor{red!10}3e-1 & \cellcolor{red!10}2e-1 & \cellcolor{red!10}4e-1 & \cellcolor{red!10}3e-1 & \cellcolor{red!10}4e-1 & \cellcolor{red!10}3e-1 \\
            \cmidrule(lr){2-11}
            & \texttt{Aar-Sampling} & \cellcolor{blue!40}7e-4 & \cellcolor{red!10}5e-1 & \cellcolor{red!10}3e-1 & \cellcolor{red!10}3e-1 & \cellcolor{blue!10}4e-3 & \cellcolor{red!10}4e-1 & \cellcolor{red!10}3e-1 & \cellcolor{red!10}4e-1 & \cellcolor{red!10}3e-1 \\
            \midrule
            \multirow{6}{*}{\centering \textbf{\textsc{Llama3-8B}}} 
            & \texttt{KGW-Logits} & \cellcolor{blue!10}3e-2 & \cellcolor{red!10}5e-1 & \cellcolor{red!10}4e-1 & \cellcolor{red!10}1e-1 & \cellcolor{red!10}8e-2 & \cellcolor{red!10}3e-1 & \cellcolor{red!10}1e-1 & \cellcolor{red!10}3e-1 & \cellcolor{red!10}2e-1 \\
            \cmidrule(lr){2-11}
            & \texttt{Aar-Logits} & \cellcolor{red!10}1e-1 & \cellcolor{red!10}5e-1 & \cellcolor{red!10}4e-1 & \cellcolor{red!10}3e-1 & \cellcolor{red!10}2e-1 & \cellcolor{red!10}3e-1 & \cellcolor{red!10}3e-1 & \cellcolor{red!10}4e-1 & \cellcolor{red!10}3e-1 \\
            \cmidrule(lr){2-11}
            & \texttt{KGW-Sampling} & \cellcolor{blue!10}5e-2 & \cellcolor{red!10}5e-1 & \cellcolor{red!10}4e-1 & \cellcolor{red!10}2e-1 & \cellcolor{red!10}1e-1 & \cellcolor{red!10}3e-1 & \cellcolor{red!10}2e-1 & \cellcolor{red!10}4e-1 & \cellcolor{red!10}2e-1 \\
            \cmidrule(lr){2-11}
            & \texttt{Aar-Sampling} & \cellcolor{red!10}3e-1 & \cellcolor{red!10}5e-1 & \cellcolor{red!10}4e-1 & \cellcolor{red!10}4e-1 & \cellcolor{red!10}3e-1 & \cellcolor{red!10}4e-1 & \cellcolor{red!10}3e-1 & \cellcolor{red!10}4e-1 & \cellcolor{red!10}3e-1 \\
            \bottomrule
        \end{tabular}
    }
    \caption{The $p$-value significance of watermarking methods under GPT-3.5 rewritten metrics, including the unmodified $p$-value, as well as the $p$-value significance after further continual pre-training, instruction tuning, DPO, and RLHF optimization. We use \raisebox{0.5ex}{\colorbox{blue!40}{\quad}} to indicate significant watermark ($p$-value < $1e-3$), \raisebox{0.5ex}{\colorbox{blue!10}{\quad}} to indicate possible watermark ($p$-value between $1e-3$ and $5e-2$), and \raisebox{0.5ex}{\colorbox{red!10}{\quad}} to indicate no watermark ($p$-value > $5e-2$).}
    \label{tab:rewrite-fine-tuning}
\end{table*}

\section{Threat Models for Backdoor-based Watermark}
\label{sec:threat2}

In~\cref{sec:rq1}, we have demonstrated that backdoor-based watermarking is an effective method for IP infringement detection and is robust against continual fine-tuning of LLMs.

The fine-tuning methods discussed in~\cref{tab:fine-tuning} assume that users are completely unaware of the trigger's existence. Under this assumption, removing the backdoor through fine-tuning is very difficult. However, removing the backdoor watermark becomes easy if users somehow become aware of the specific trigger. Therefore, future research should focus on making the trigger as undetectable as possible (even rare word combinations are at risk of being discovered) and on verifying backdoor watermarks without exposing the trigger.

\section{Detailed Accuracy Reference for \texorpdfstring{\cref{tab:fine-tuning}}{Table 1}}
\label{appendix:accuracy}

In~\cref{tab:fine-tuning}, we only show the $p$-value metrics for the Open-Source LLM Intellectual Property Detection scenario. For reference,~\cref{tab:acc-fine-tuning} provides the corresponding original accuracy for each $p$-value. The backdoor method indicates the correct trigger rate, while the inference-time watermark distillation method refers to the accuracy of watermark and human text at a $p$-value of 0.05, as described in~\cref{sec:distillation-detection1}.

\begin{table*}[t]
    \small
    \centering
    \resizebox{1 \textwidth}{!}{%
        \begin{tabular}{lcccccccccc}
            \toprule
            & & & \multicolumn{2}{c}{\textit{\textbf{W. Continual PT}}} & \multicolumn{2}{c}{\textit{\textbf{W. Continual IT}}} & \multicolumn{2}{c}{\textit{\textbf{W. Continual IT+DPO}}} & \multicolumn{2}{c}{\textit{\textbf{W. Continual IT+RLHF}}} \\
            \cmidrule(lr){4-5} \cmidrule(lr){6-7} \cmidrule(lr){8-9} \cmidrule(lr){10-11}
            \multirow{-2}{*}{\centering \textbf{Target-LLM}} & \multirow{-2}{*}{\centering \textbf{\makecell{Watermark\\Methods}}} & \multirow{-2}{*}{\centering \textbf{\makecell{$P$-Value$\downarrow$\\(Origin)}}} & \textbf{Full$\downarrow$} & \textbf{LoRA$\downarrow$} & \textbf{Full$\downarrow$} & \textbf{LoRA$\downarrow$} & \textbf{Full$\downarrow$} & \textbf{LoRA} & \textbf{Full$\downarrow$} & \textbf{LoRA$\downarrow$} \\
            \midrule
            \rowcolor{gray!25}
            \multicolumn{11}{c}{\textbf{Scenario 1: Open-Source LLM Intellectual Property Detection} (\cref{sec:scenarios-model})} \\
            \midrule
            \multirow{8}{*}{\centering \textbf{\textsc{Llama2-7B}}} 
            & \texttt{Backdoor-PT} & 33.0\% & 18.0\% & 22.0\% & 33.5\% & 32.5\% & 30.5\% & 31.5\% & 33.0\% &32.5\%  \\
            \cmidrule(lr){2-11}
            & \texttt{Backdoor-IT} & 34.0\% & \textbf{N/A} & \textbf{N/A} & 28.5\% & 29.5\% & 29.0\% & 30.0\%  & 30.5\% & 31.5\%  \\
            \cmidrule(lr){2-11}
            & \texttt{KGW-Logits} &  98.0\% & 53.4\% & 68.5\% & 89.9\% &  94.1\% & 85.2\%  & 91.9\% & 84.6\% & 90.5\%  \\
            \cmidrule(lr){2-11}
            & \texttt{Aar-Logits} & 96.5\% & 50.4\% & 56.3\% & 79.4\% & 90.9\% & 81.4\% & 89.3\% & 80.3\% & 87.9\%   \\
            \cmidrule(lr){2-11}
            & \texttt{KGW-Sampling} & 94.4\% & 53.3\% & 61.3\% & 74.0\% & 92.9\% & 70.8\% & 73.5\% & 83.4\% & 87.9\% \\
            \cmidrule(lr){2-11}
            & \texttt{Aar-Sampling} & 95.5\% & 50.28\% & 56.9\% & 73.4\% & 94.9\% & 61.1\% & 77.0\% & 78.9\% & 83.4\% \\
            \midrule
            \multirow{8}{*}{\centering \textbf{\textsc{Llama3-8B}}} 
            &  \texttt{Backdoor-PT} & 82.5\%&	33.5\%&	42.5\%&	75.5\%	&78.5\%&	74.0\%	&77.0\%&	75.5\%&	80.5\%  \\
            \cmidrule(lr){2-11}
            &  \texttt{Backdoor-SFT} & 83.5\% & \textbf{N/A} & \textbf{N/A} & 80.5\%	&82.5\%	&79.5\%	&80.0\%&	79.0\%	&81.5\% \\
            \cmidrule(lr){2-11}
            & \texttt{KGW-Logits} & 98.5\% & 53.4\% & 71.3\%  & 84.9\% & 90.8\% & 80.1\%  & 89.7\% & 79.8\%  & 88.7\%  \\
            \cmidrule(lr){2-11}
            & \texttt{Aar-Logits} & 98.5\% & 51.4\% & 57.4\%& 81.4\%  & 87.3\$ & 78.9\%  & 86.5\% & 78.0\%  & 85.7\%  \\
            \cmidrule(lr){2-11}
            & \texttt{KGW-Sampling} & 97.8\% & 52.3\% & 65.1\% & 81.0\% & 84.6\% & 77.8\% & 87.6\% &  78.6\%  & 86.7\%  \\
            \cmidrule(lr){2-11}
            & \texttt{Aar-Sampling} & 96.5\% & 50.6\% & 55.2\% & 72.9\% & 81.4\% & 73.1\%  & 81.3\%  & 72.9\%  &  80.7\%  \\
            \bottomrule
        \end{tabular}
    }
    \caption{The accuracy of watermarking methods under two scenarios, including the unmodified accuracy, as well as the accuracy after further continual pre-training, instruction tuning, DPO, and RLHF optimization.}
    \label{tab:acc-fine-tuning}
\end{table*}

\section{Dataset Descriptions}
This section presents the datasets used in our experiments, detailing their names, sizes, sources, and key characteristics. The datasets are categorized based on the capabilities they assess, including understanding and reasoning capabilities, generation capability, continual pre-training, supervised instruct tuning, and alignment optimization. The descriptions are summarized in~\cref{tab:dataset_descriptions}.
\begin{table*}[ht]
    \centering
    \setlength{\tabcolsep}{10pt} 
    \renewcommand{\arraystretch}{1.2} 
    \resizebox{\textwidth}{!}{
    \begin{tabular}{@{}lccl@{}}
    \toprule
    \textbf{Dataset} & \textbf{Size} & \textbf{Source} & \textbf{Key Characteristics} \\ \midrule
    \multicolumn{4}{c}{\textbf{Understanding and Reasoning Capabilities}} \\ \midrule
    \textbf{\textsc{Arc-Easy}}~\cite{clark2018think} & 5,197 & AI2 & Elementary science questions\\[0.5em]
    \textbf{\textsc{Arc-Challenge}}~\cite{clark2018think} & 2,590 & AI2 & More difficult science questions \\[0.5em]
    \textbf{\textsc{HellaSwag}}~\cite{zellers2019hellaswag} & 59.950 & Web & Commonsense reasoning tasks \\[0.5em]
    \textbf{\textsc{MMLU}}~\cite{hendrycks2020measuring} & 231,400 & Diverse & Multi-domain, multiple-choice questions  \\[0.5em]
    \textbf{\textsc{Winogrande}}~\cite{sakaguchi2021winogrande} & $\sim$44K & Crowd-sourced & Pronoun resolution challenges \\ \midrule
    \multicolumn{4}{c}{\textbf{Generation Capability}} \\ \midrule
    \textbf{\textsc{WikiText}}~\cite{merity2016pointer} & 3,708,608 & Wikipedia & High-quality articles from Wikipedia \\ \midrule
    \multicolumn{4}{c}{\textbf{Continual Pre-training}} \\ \midrule
    \textbf{\textsc{C4}}~\cite{raffel2020exploring} & $\sim$10B & Web & Massive, clean crawled corpus\\ \midrule
    \multicolumn{4}{c}{\textbf{Supervised Instruct Tuning}} \\ \midrule
    \textbf{\textsc{Alpaca}}~\cite{alpaca} & $\sim$52K & Self-generated & Instruction-following dataset from GPT-4 \\ \midrule
    \multicolumn{4}{c}{\textbf{Alignment Optimization}} \\ \midrule
    \textbf{\textsc{HH-RLHF}}~\cite{bai2022training} & 169,352 & Human Feedback & Human preference data for RLHF \\ \bottomrule
    \end{tabular}
    }
    \caption{Dataset Descriptions Across Multiple Capabilities}
    \label{tab:dataset_descriptions}
    \end{table*}

\section{Details of continual fine-tuning Method}
\label{sec:further-fine-tuning}

This section details the various continual fine-tuning methods used in this work, including continual pre-training, continual instruction tuning, continual direct preference learning, and continual reinforcement learning from human feedback. We also discuss the use of the LoRA fine-tuning approach.

\subsection{Continual Pre-training}
Continual pre-training involves unsupervised training of the language model on a large corpus. Using text data $\{x\}$, the objective is to maximize the probability of the next token:
\begin{equation}
\mathcal{L}_{PT}=-\sum_{i=1}^{n}\log P(x_i|x_{<i};M),
\end{equation}
where $M$ are the model parameters and $n$ is the length of the text $x$. This autoregressive training helps the model learn the statistical features of the text.

\subsection{Continual Instruction Tuning}
Continual instruction tuning builds on pre-training by using an instruction dataset $\{(x,y)\}$ to fine-tune the model so it can generate appropriate answers to given instructions. Here, $x$ is the instruction or question, and $y$ is the corresponding answer. The loss function maximizes the conditional probability:
\begin{equation}
\mathcal{L}_{IT}=-\sum_{i=1}^{m}\log P(y_i|x_i;M),
\end{equation}
where $m$ is the size of the training data. This approach teaches the model to explicitly generate results based on instructions.

\subsection{Continual Direct Preference Learning}
Continual direct preference learning (DPO)~\cite{rafailov2023direct} uses human-labeled preference data $\{(x,y_1,y_2)\}$ to learn preferences, where $y_1$ and $y_2$ are two candidate answers generated by the model for $x$. The labeled data indicates whether $y_1$ is preferred over $y_2$ or vice versa. The training minimizes the pairwise ranking loss:
\begin{align}
    \mathcal{L} = -\mathbb{E}_{(x, y_1, y_2)\sim \mathcal{D}_{pref}}\left[\log \sigma \left(\beta \log \frac{\pi_{M}(y_1\mid x)}{\pi_\text{ref}(y_1\mid x)}  - \beta \log \frac{\pi_{M}(y_2\mid x)}{\pi_\text{ref}(y_2\mid x)}\right)\right],
\end{align}
where $\sigma$ is the logistic function, $\beta$ is a scaling parameter, and $\pi_{M}$ and $\pi_\text{ref}$ are the probability distributions of the current model and the reference model, respectively.

\subsection{Continual Reinforcement Learning From Human Feedback}
Reinforcement Learning from Human Feedback (RLHF)~\cite{ouyang2022training} involves two main stages. First, a reward model is trained using a dataset of human-labeled preferences. Second, this reward model, combined with the PPO algorithm, is used to train a language model via reinforcement learning.

To train the reward model, the data includes an input $x$ and two outputs $y_w$ and $y_l$, where $y_w$ is the preferred response. The reward model, represented as $r^*(y, x)$, uses the Bradley-Terry (BT) model~\cite{bradley1952rank} to express preference probabilities:

\begin{equation}
P(y_w \succ y_l \mid x) = \frac{\exp(r^*(y_w, x))}{\exp(r^*(y_w, x)) + \exp(r^*(y_l, x))}.
\end{equation}

Here, $P(y_w \succ y_l \mid x)$ is the probability that $y_w$ is preferred over $y_l$ given $x$. The reward model $r^*(y, x)$ scores each potential output $y$. The BT model is commonly used for pairwise comparisons to represent these preferences.

Given the training data $\{x, y_w, y_l\}_i^N$, the reward model $r_M(y, x)$ is trained using the following loss function:

\begin{equation}
\mathcal{L}_R(r_M, D) = -\mathbb{E}_{(x,y_w,y_l)} \left[ \log \sigma(r_M(y_w, x) - r_M(y_l, x)) \right].
\end{equation}

Here, $\sigma$ is the logistic function, and the expectation is over triplets $(x, y_w, y_l)$ from $D$. This loss function pushes the model to score the preferred output $y_w$ higher than $y_l$ for a given $x$. Minimizing this loss enables the reward model to learn human preferences.

In the reinforcement learning phase, the trained reward model guides the language model training. The aim is to optimize the language model's policy $\pi_M$ to maximize the expected reward from $r_M$, while keeping outputs close to a reference policy $\pi_{\text{ref}}$. This is achieved with the following objective:

\begin{equation}
\max_{\pi_M} \mathbb{E}_{x, y \sim \pi_M} \left[ r_M(y, x) \right] - \beta \mathbb{D}_{\text{KL}} \left[ \pi_M \parallel \pi_{\text{ref}} \right].
\end{equation}

This balances enhancing the language model's performance and maintaining alignment with human preferences.

\subsection{Low Rank Adaptation (LoRA) Fine-tuning Method}

We utilize both full-parameter tuning and LoRA (Low-Rank Adaptation) fine-tuning for continual fine-tuning. The core idea of LoRA fine-tuning is to adjust only the low-rank projection matrices while keeping the other parameters of the pre-trained model fixed.

Assume the weight matrix of the original model is $\mathbf{W} \in \mathbb{R}^{d \times k}$. LoRA defines two low-rank matrices $\mathbf{A} \in \mathbb{R}^{d \times r}$ and $\mathbf{B} \in \mathbb{R}^{r \times k}$, where $r \ll \min(d, k)$. The augmented weight matrix is:

\begin{equation}
\mathbf{W}_{\text{lora}} = \mathbf{W} + \mathbf{A}\mathbf{B}.
\end{equation}

During fine-tuning, only matrices $\mathbf{A}$ and $\mathbf{B}$ are updated, while $\mathbf{W}$ remains unchanged. This method retains the knowledge of the original model, significantly reduces the number of parameters to be fine-tuned, and speeds up the training process. Due to its lower resource requirements, LoRA is often preferred for fine-tuning open-source models.

\section{Details of Generation Capability Metrics}
\label{appendix:generation-metrics}

To evaluate the generation capability of the fine-tuned models, we employed two metrics: Perplexity (PPL) and Sequence Repetition of 3-grams (Seq-Rep-3), computed on the \textsc{WikiText} dataset~\cite{merity2016pointer}.

\noindent \textbf{Perplexity (PPL)} measures the fluency and coherence of the generated text. A lower PPL indicates better predictive capability and higher quality. Specifically, we calculate PPL using the \textsc{Llama2-70B}~\cite{llama3modelcard} model to assess the language model’s effectiveness in predicting tokens. The calculation of PPL is defined as follows:

\begin{equation}
\text{PPL}(x) = \exp\left(-\frac{1}{n}\sum_{i=1}^{n}\log P(x_i \mid x_{<i}; M)\right),
\end{equation}

where $x = (x_1, x_2, \dots, x_n)$ is the sequence, and M denotes the language model parameters. To ensure a fair assessment of fluency rather than benefiting from repetitive text, we set a no-repeat $n$-gram constraint $(n=5)$ during evaluation. This setting prevents artificially low PPL scores due to repetition.

\noindent \textbf{Sequence Repetition (Seq-Rep-3)} evaluates text diversity by measuring the proportion of repeated 3-grams in generated sequences~\cite{welleck2019neural}. A lower Seq-Rep-3 indicates more diverse and natural text generation. Formally, Seq-Rep-3 is defined as:

\begin{equation}
\text{Seq-Rep-3}(x) = \frac{\text{number of repeated 3-grams in sequence } x}{\text{total number of 3-grams in sequence } x}.
\end{equation}

These metrics provide complementary insights into the quality and diversity of model-generated text. PPL measures fluency, with lower values indicating better token prediction and coherence. However, low PPL alone may not reflect text diversity, as repetition can artificially reduce it. Seq-Rep-3 addresses this by quantifying 3-gram repetitions, with lower values indicating more varied text.

\section{License Overview of Various Open-source LLMs}
\label{appendix:A}

This appendix provides an overview of the licensing terms for several open-source Large Language Models (LLMs), highlighting their alignments with our scenarios.

\subsection{Meta Llama Series}
Meta's use policy for Llama 2\footnote{\url{https://ai.meta.com/llama/use-policy/}} outlines several prohibited uses, which directly relate to the scenarios defined in our paper.

\paragraph{Scenario 1: Detecting IP Infringement}

\begin{itemize}[label={}]
    \item 
    \refstepcounter{table}
    \begin{tcolorbox}[
        standard jigsaw,
        title=Policy 1.g,
        opacityback=0,
    ]
    Engage in or facilitate any action or generate any content that infringes, misappropriates, or otherwise violates any third-party rights, including the outputs or results of any products or services using the Llama 2 Materials.
    \end{tcolorbox}
    - This policy directly relates to IP Detection, as it prohibits actions that infringe on intellectual property rights. The scenario's goal of detecting unauthorized commercial use or copying of open-source LLMs is supported by this clause.
\end{itemize}

\paragraph{Scenario 2: Detecting Generated Text}

\begin{itemize}[label={}]
    \item 
    \refstepcounter{table}
    \begin{tcolorbox}[
        standard jigsaw,
        title=Policy 1.a,
        opacityback=0,
    ]
    ``Engage in, promote, generate, contribute to, encourage, plan, incite, or further illegal or unlawful activity or content, such as:
    \begin{itemize}
\item Violence or terrorism
\item Exploitation or harm to children, including the solicitation, creation, acquisition, or dissemination of child exploitative content or failure to report Child Sexual Abuse Material''
\end{itemize}
    \end{tcolorbox}
    - This policy is pertinent to Generated Text Detection, which aims to detect whether generated text from an open-source LLM contains illegal, harmful, or unethical content. The prohibition of generating such content aligns with the scenario's goal of preventing misuse for disseminating harmful material.

    \item 
    \refstepcounter{table}
    \begin{tcolorbox}[
        standard jigsaw,
        title=Policy 1.c,
        opacityback=0,
    ]
    ``Engage in, promote, incite, or facilitate the harassment, abuse, threatening, or bullying of individuals or groups of individuals.''
    \end{tcolorbox}
    - This policy relates to Generated Text Detection by prohibiting the generation of abusive or harassing content. The scenario's goal of detecting harmful generated text includes identifying text that facilitates harassment or abuse.

    \item 
    \refstepcounter{table}
    \begin{tcolorbox}[
        standard jigsaw,
        title=Policy 3.a,
        opacityback=0,
    ]
    ``Intentionally deceive or mislead others, including use of Llama 2 related to the following:
    \begin{itemize}
\item Generating, promoting, or furthering fraud or the creation or promotion of disinformation''
\end{itemize}
    \end{tcolorbox}
    - This policy supports Generated Text Detection by addressing the misuse of LLMs to generate misleading or fraudulent content. Detecting such generated text aligns with the policy's goal of preventing deception and misinformation.

    \item 
    \refstepcounter{table}
    \begin{tcolorbox}[
        standard jigsaw,
        title=Policy 3.e,
        opacityback=0,
    ]
    ``Representing that the use of Llama 2 or outputs are human-generated.''
    \end{tcolorbox}
    - This policy underlines the importance of transparency in content generation. Generated Text Detection's goal is to determine whether a text is generated by an LLM or its fine-tuned version aligns with ensuring users do not misrepresent AI-generated content as human-generated.
\end{itemize}

In summary, The Llama 2 use policy explicitly prohibits various activities related to both scenarios defined in our paper, particularly IP infringement and the generation of illegal or harmful content. 

\subsection{Command R Series}
Cohere R series is built on the language of business and is optimized for enterprise generative AI, search and discovery, and advanced retrieval. Their Cohere For AI Acceptable Use Policy\footnote{\url{https://docs.cohere.com/docs/c4ai-acceptable-use-policy}} aligns with our scenario settings.
\paragraph{Scenario 1: Detecting IP Infringement}

\begin{itemize}[label={}]
    \item \refstepcounter{table}
    \begin{tcolorbox}[
        standard jigsaw,
        title=Cohere For AI Acceptable Use Policy,
        opacityback=0,
    ]
    ``Synthetic data for commercial uses: generating synthetic data outputs for commercial purposes, including to train, improve, benchmark, enhance or otherwise develop model derivatives, or any products or services in connection with the foregoing.''
    \end{tcolorbox}
    - This policy is highly relevant to IP Detection, as it explicitly prohibits using models or their derivatives for commercial purposes without permission, which is a core concern of detecting IP infringement.
\end{itemize}

\paragraph{Scenario 2: Detecting Generated Text}

\begin{itemize}[label={}]
    \item 
    \refstepcounter{table}
    \begin{tcolorbox}[
        standard jigsaw,
        title=Cohere For AI Acceptable Use Policy,
        opacityback=0,
    ]
    ``We expect users of our models or model derivatives to comply with all applicable local and international laws and regulations. Additionally, you may not use or allow others to use our models or model derivatives in connection with any of the following strictly prohibited use cases:''
    \end{tcolorbox}
    - This policy establishes the baseline expectation that users must comply with all laws and regulations, which supports the detection of misuse in both scenarios. Generated Text Detection specifically aligns with preventing the generation and dissemination of illegal content.

    \item \refstepcounter{table}
    \begin{tcolorbox}[
        standard jigsaw,
        title=Cohere For AI Acceptable Use Policy,
        opacityback=0,
    ]
    ``Harassment and abuse: engaging in, promoting, facilitating, or inciting activities that harass or abuse individuals or groups.''
    \end{tcolorbox}
    - This policy supports Generated Text Detection by setting clear boundaries against generating content that could harass or abuse individuals or groups, aligning with the scenario's goals of detecting unethical content.

    \item \refstepcounter{table}
    \begin{tcolorbox}[
        standard jigsaw,
        title=Cohere For AI Acceptable Use Policy,
        opacityback=0,
    ]
    ``Violence and harm: engaging in, promoting, or inciting violence, threats, hate speech self-harm, sexual exploitation, or targeting of individuals based on protected characteristics.''
    \end{tcolorbox}
    - This policy directly relates to Generated Text Detection, where the goal is to detect generated content that disseminates illegal, harmful, or unethical content. It provides a clear mandate against such misuse.

    \item \refstepcounter{table}
    \begin{tcolorbox}[
        standard jigsaw,
        title=Cohere For AI Acceptable Use Policy,
        opacityback=0,
    ]
    ``Fraud and deception: misrepresenting generated content from models as human-created or allowing individuals to create false identities for malicious purposes, deception, or to cause harm, through methods including:
    \begin{itemize}
    \item propagation of spam, fraudulent activities such as catfishing, phishing, or generation of false reviews;
    \item creation or promotion of false representations of or defamatory content about real people, such as deepfakes; or
    \item creation or promotion of intentionally false claims or misinformation.''
    \end{itemize}
    \end{tcolorbox}
    - This is pertinent to both scenarios. For IP Detection, it addresses the misrepresentation of generated content as human-created, which can involve claiming an open-source LLM as a proprietary creation. For Generated Text Detection, it covers the generation of harmful or deceptive content.
\end{itemize}

\subsection{01.ai Yi Series}
The Yi series is another open-source LLM that has demonstrated excellent performance on the LMSYS Chatbot Arena Leaderboard\footnote{\url{https://arena.lmsys.org/}}. This series of LLMs has been developed by a Chinese company named ``Lingyiwanwu.'' The following user agreement\footnote{\url{https://platform.lingyiwanwu.com/useragreement}} contains sections that align with the assumptions in our scenarios.

\paragraph{Scenario 1: Detecting IP Infringement}

\begin{itemize}[label={}]
    \item 
    \refstepcounter{table}
    \begin{tcolorbox}[
        standard jigsaw,
        title=Article 5\, Clause 1,
        opacityback=0,
    ]
    ``Lingyiwanwu is the developer and operator of this product and enjoys all rights to the data, information, and outputs generated during the development and operation of this product within the scope permitted by laws and regulations, except where the relevant rights holders are entitled to rights according to law.''
    \end{tcolorbox}
    - This clause asserts that Lingyiwanwu holds the rights to the outputs generated by the product, reinforcing the need to detect IP infringement when these rights are violated by unauthorized use or copying of the LLMs.

    \item 
    \refstepcounter{table}
    \begin{tcolorbox}[
        standard jigsaw,
        title=Article 5\, Clause 2,
        opacityback=0,
    ]
    ``Unless otherwise agreed or stipulated by laws and regulations, you have the rights to the content generated based on the content you are entitled to upload and the rights to the content generated based on the uploaded content.''
    \end{tcolorbox}
    - This clause delineates user rights to generated content, provided it is based on legally uploaded content, highlighting the importance of detecting if generated content infringes on existing IP rights.

    \item 
    \refstepcounter{table}
    \begin{tcolorbox}[
        standard jigsaw,
        title=Article 5\, Clause 6,
        opacityback=0,
    ]
    ``You understand and promise that your input during the use of this product will not infringe on any person's intellectual property rights, portrait rights, reputation rights, honor rights, name rights, privacy rights, personal information rights, etc. Otherwise, you will bear the risk and responsibility of infringement.''
    \end{tcolorbox}
    - This clause ensures that users acknowledge their responsibility to avoid infringing on IP rights, aligning with the scenario’s assumption that detection mechanisms are needed to prevent such infringements.

    \item 
    \refstepcounter{table}
    \begin{tcolorbox}[
        standard jigsaw,
        title=Article 5\, Clause 7,
        opacityback=0,
    ]
    ``If you add new data for model training, fine-tuning, and development during the use of this product, you will bear the resulting responsibilities.''
    \end{tcolorbox}
    - This clause emphasizes user responsibility for any new data added for model training or fine-tuning, aligning with the scenario’s focus on detecting whether the generated text has been modified or fine-tuned from the original LLM.

\end{itemize}

\paragraph{Scenario 2: Detecting Generated Text}

\begin{itemize}[label={}]
    \item 
    \refstepcounter{table}
    \begin{tcolorbox}[
        standard jigsaw,
        title=Article 4\, Clause 1,
        opacityback=0,
    ]
    ``Based on your use of this product, Lingyiwanwu grants you a revocable, non-transferable, non-exclusive right to use this product. If you publish or disseminate content generated by this product, you should: 
    \begin{itemize}
        \item Proactively verify the authenticity and accuracy of the output content to avoid spreading false information;
        \item Mark the output content as AI-generated in a prominent way to inform the public about the content synthesis;
        \item Avoid publishing and disseminating any output content that violates the usage norms of this agreement.''
    \end{itemize}
    \end{tcolorbox}
    - This clause mandates users to verify and label AI-generated content, ensuring transparency and preventing the misuse of generated text for harmful or illegal purposes, which aligns with the scenario’s goal of detecting and managing generated content responsibly.

    \item 
    \refstepcounter{table}
    \begin{tcolorbox}[
        standard jigsaw,
        title=Article 4\, Clause 4,
        opacityback=0,
    ]
    ``Users are prohibited from engaging in certain behaviors, including but not limited to:
    \begin{itemize}
        \item \textbf{(5)} Inducing the generation of content that violates relevant laws and regulations or contains unfriendly outputs;
        \item \textbf{(7)} Developing products and services that compete with this product using this product;
        \item \textbf{(9)} Unauthorized removal or alteration of AI-generated labels or deep synthesis content labels.''
    \end{itemize}
    \end{tcolorbox}
    - These prohibitions directly support the scenario's assumptions by preventing the generation and dissemination of harmful content, ensuring ethical use of the model, and maintaining the integrity of AI-generated labels for accountability.

\end{itemize}

\section{System Specifications for Reproductivity}
\label{appendix:specs}
Our experiments were conducted on high-performance servers, each equipped with either an Intel(R) Xeon(R) Platinum 8378A CPU @ 3.00GHz or an Intel(R) Xeon(R) Platinum 8358P CPU @ 2.60GHz, 1TB of RAM, and 4/6/8 NVIDIA A800 GPUs with 80GB memory. Machines with 4/8 GPUs are configured with the SXM4 version, while those with 6 GPUs use the PCIe version. The software environment included \textit{Python} 3.11, \textit{PyTorch} 2.3, and \textit{NCCL} 2.20.5 for reproductivity.

\end{document}